\pdfoutput=1

\documentclass[11pt]{article}

\usepackage[preprint]{acl}

\usepackage{times}
\usepackage{latexsym}

\usepackage[T1]{fontenc}

\usepackage[utf8]{inputenc}

\usepackage{microtype}

\usepackage{inconsolata}
\usepackage{makecell}
\usepackage{tabu}
\usepackage{multicol}
\usepackage[utf8]{inputenc} %
\usepackage[T1]{fontenc}    %
\usepackage{hyperref}       %
\usepackage{url}            %
\usepackage{booktabs}       %
\usepackage{amsfonts}       %
\usepackage{nicefrac}       %
\usepackage{microtype}      %
\usepackage{xcolor}         %
\usepackage{microtype}
\usepackage{hyperref}
\usepackage{url}
\usepackage{colortbl}
\usepackage{wrapfig}
\usepackage{booktabs}
\definecolor{darkblue}{rgb}{0, 0, 0.5}
\usepackage{algorithm}%
\usepackage{algpseudocode}

\usepackage{amsmath,amssymb}
\hypersetup{colorlinks=true, citecolor=darkblue, linkcolor=darkblue, urlcolor=darkblue}

\usepackage{graphicx} %
\usepackage{multirow}
\usepackage{pifont}
\usepackage{bm} %
\usepackage{amsmath} %
\usepackage{tabularx}
\usepackage{longtable}
\usepackage{booktabs}
\usepackage{arydshln}
\usepackage{wrapfig}
\usepackage{graphicx} %
\usepackage{subcaption} %

\usepackage{amsmath,amsfonts,bm}

\def\eqref#1{equation~\ref{#1}}

\def\1{\bm{1}}

\def\vs{{\bm{s}}}

\def\mE{{\bm{E}}}

\def\mH{{\bm{H}}}
\def\mI{{\bm{I}}}

\def\mW{{\bm{W}}}
\def\mX{{\bm{X}}}

\DeclareMathAlphabet{\mathsfit}{\encodingdefault}{\sfdefault}{m}{sl}
\SetMathAlphabet{\mathsfit}{bold}{\encodingdefault}{\sfdefault}{bx}{n}

\DeclareMathOperator*{\argmin}{arg\,min}

\newcommand{\red}[1]{\textcolor{red}{#1}}

\newcommand{\gr}[1]{\textcolor{gray}{#1}}

\newcommand{\mg}[1]{\textcolor{magenta}{#1}}
\newcommand{\pp}[1]{\textcolor{violet}{#1}}
\newcommand{\og}[1]{\textcolor{orange}{#1}}

\usepackage{placeins}
\usepackage{tcolorbox}
\usepackage{listings}
\usepackage{enumitem}
\usepackage{adjustbox}

\usepackage{subcaption}

\title{LLMC: Benchmarking Large Language Model Quantization with a Versatile Compression Toolkit}

\newcommand*\samethanks[1][\value{footnote}]{\footnotemark[#1]}
\author{Ruihao Gong\hspace{4pt}$^{1,2}$\thanks{Equal contribution.} \quad
Yang Yong\hspace{4pt}$^2$\samethanks \quad
Shiqiao Gu\hspace{4pt}$^2$\samethanks \quad
Yushi Huang\hspace{4pt}$^{1,2}$\samethanks \quad \\
\textbf{Chengtao Lv}$^{1,2}$ \quad
\textbf{Yunchen Zhang}$^2$ \quad
\textbf{Dacheng Tao}$^3$ \quad
\textbf{Xianglong Liu}\hspace{4pt}$^1$\thanks{Corresponding authors.}\quad \\
$^1$Beihang University \quad $^2$SenseTime Research \quad $^3$Nanyang Technological University \\
\normalsize\texttt{\{gongruihao, yongyang, gushiqiao, huangyushi, lvchengtao, zhangyunchen\}@sensetime.com}\\\normalsize\quad\texttt{xlliu@buaa.edu.cn}\quad\texttt{dacheng.tao@ntu.edu.sg}
}

\begin{document}
\maketitle
\begin{abstract}
Recent advancements in large language models (LLMs) are propelling us toward artificial general intelligence with their remarkable emergent abilities and reasoning capabilities. However, the substantial computational and memory requirements limit the widespread adoption. Quantization, a key compression technique, can effectively mitigate these demands by compressing and accelerating LLMs, albeit with potential risks to accuracy. Numerous studies have aimed to minimize the accuracy loss associated with quantization. However, their quantization configurations vary from each other and cannot be fairly compared. In this paper, we present LLMC, a plug-and-play compression toolkit, to fairly and systematically explore the impact of quantization. LLMC integrates dozens of algorithms, models, and hardware, offering high extensibility from integer to floating-point quantization, from LLM to vision-language (VLM) model, from fixed-bit to mixed precision, and from quantization to sparsification. Powered by this versatile toolkit, our benchmark covers three key aspects: calibration data, algorithms (three strategies), and data formats, providing novel insights and detailed analyses for further research and practical guidance for users. Our toolkit is available at \url{https://github.com/ModelTC/llmc}.
\end{abstract}

\section{Introduction}
Recently, LLMs such as GPT-4~\citep{openai2024gpt4} have demonstrated unprecedented generative capabilities in the field of natural language processing~(NLP) and also achieved widespread applications. However, their substantial computational and storage costs have impeded their further popularization among users. For instance, BLOOM~\citep{touvron2023llama}, a multilingual LLM with 176 billion parameters, requires a minimum of 350 GB space to store model weights in full-precision~(FP16) format. Even worse, it requires 5$\times$80GB A100 or 9$\times$40GB A800 NVIDIA GPUs to perform inference. Therefore, reducing LLMs' serving cost is paramount to further enhance their application.

For the aforementioned challenge, model quantization~\citep{nagel2021white} can be an effective solution. It maps weights and/or activations to a lower-bit data format to reduce memory footprint and accelerate model inference. Existing quantization approaches can be categorized into two types: quantization-aware-training~(QAT)~\citep{bhalgat2020lsq, 2019-iccv-dsq, esser2020learned} and post-training quantization~(PTQ)~\citep{wei2023qdrop, li2021brecq}. Although with prominent high performance, the necessity for QAT to undergo finetuning or retraining with substantial training data and training costs renders it unattainable for the majority of users. Correspondingly, PTQ compresses models without retraining, making it a preferred method for LLMs due to its minimal resource requirements. Therefore, we do not mention some QAT methods~\citep {du2024bitdistiller, liu2024qllm, liu2023llmqat, egiazarian2024extreme} in this paper. 

\begin{figure*}[ht!]
\vspace{-0.2in}
   \centering
       \includegraphics[width=0.98\linewidth]{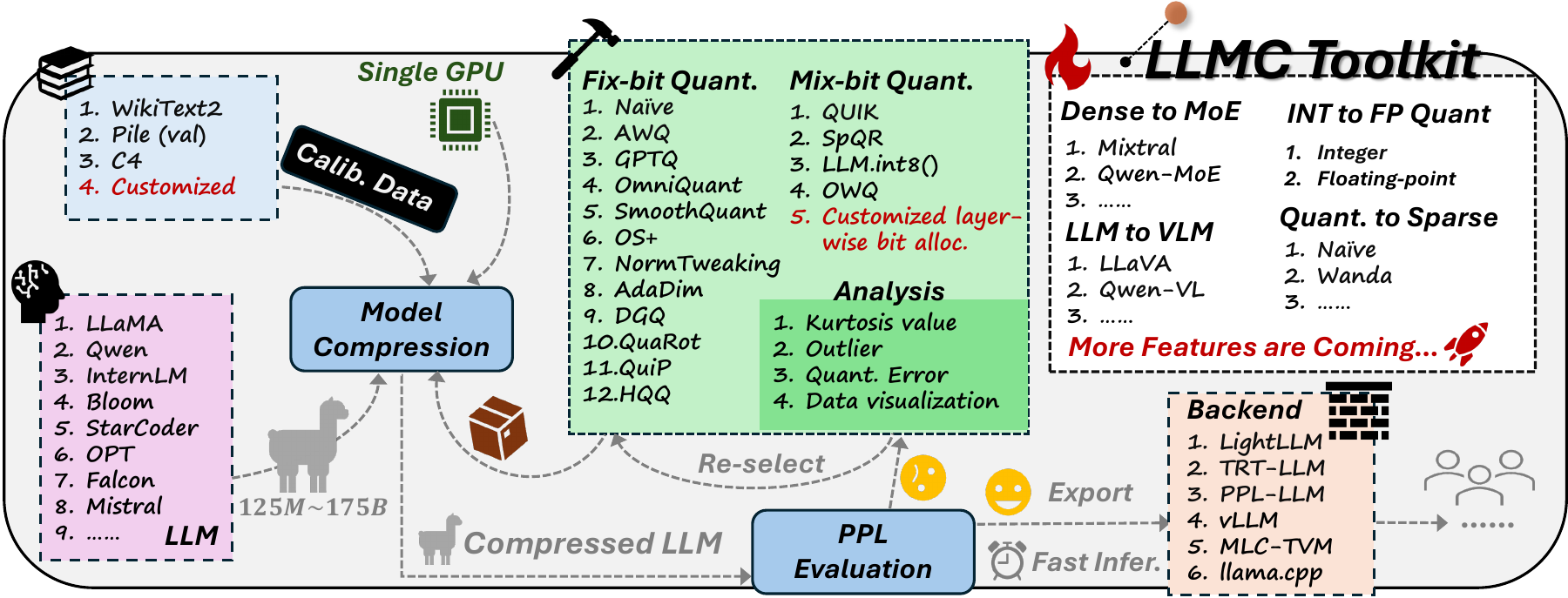}
        \vspace{-0.05in}
       \caption{Overview of our LLM compression toolkit LLMC, which incorporates diverse algorithms, ultra-low cost quantization, multiple backends support, and high extensibility. More features are under development.} %
   \label{fig:overview}
   \vspace{-0.2in}
\end{figure*}
However, current PTQ methods always evaluate across distinct datasets in different quantization configurations and with simulated quantization. For example, AWQ~\cite{lin2023awq} employs Pile (val)~\cite{pile} as calibration data, instead of C4~\cite{2019t5} in GPTQ~\cite{frantar2022gptq}. This situation would cause an inaccurate assessment of configurations for efficient and accurate LLM quantization. 

To provide a comprehensive options menu for users and directions with insights for further research, we make a fair benchmark, which considers three key dimensions, \emph{e.g.}, calibration data, algorithms, and data formats. First, we systematically explore the effect of calibration data for higher model performance. Then, we aim to investigate the effectiveness and underlying mechanisms of three primary algorithm strategies: transformation, clipping, and reconstruction. Finally, we probe how to select types between the integer and float-point quantization for further accuracy improvements. All the aforementioned studies benefit from our LLMC, a user-friendly, plug-and-play LLM compression toolkit. This toolkit incorporates several distinct traits, as demonstrated in \autoref{fig:overview}, offering users the freedom to select options that best suit their needs. 

In a word, our main contributions can be described as follows: 
\begin{itemize}[leftmargin=*]
    \item We release a versatile LLM compression toolkit LLMC supporting dozens of algorithms, models, and multiple inference backends with powerful expandability and all-around evaluation. It also enables users to perform compression for 100-billion-parameter LLMs with just a single GPU, which substantially facilitates the application of LLM quantization.
    \item We modularly and fairly benchmark LLM quantization considering calibration data, algorithms, and data formats. With detailed observation and analysis, we provide various types of novel points for performance and method improvements under different configurations.
    \item Equipped with our powerful toolkit and comprehensive insights, future LLM researchers can efficiently integrate suitable algorithms and low-bit formats for their applications, thereby democratizing the compression of large language models.
\end{itemize}

\section{LLMC: A Versatile LLM Compression Toolkit}
First and foremost, we have developed a comprehensive toolkit named LLMC for LLM compression, characterized by the following key features, which are also exhibited in \autoref{fig:overview}.

\noindent\textbf{Diverse algorithms support.} LLMC supports a wide range of quantization algorithms, including 16 different methods covering weight-only, weight-activation, and mixed-precision quantization. This variety allows for fair comparisons and in-depth analyses of different approaches.

\noindent\textbf{Quantization with an ultra-low cost.} Our toolkit is designed to be resource-efficient, and capable of running large models with minimal hardware requirements. Benefiting from our pipeline with offloading technique, only one 40G A100 is required to calibrate and evaluate OPT-175B~\cite{zhang2022optopenpretrainedtransformer}, whose weights occupies $\approx$ 350GB.%

\noindent\textbf{Multi-backend compatibility.} Built on LLMC, various quantization settings and model formats are compatible with multiple backends and hardware platforms, such as LightLLM~\cite{lightllm}, TRT-LLM~\cite{trt-llm}, PPL-LLM~\cite{ppl-llm}, vLLM~\cite{kwon2023efficient}, MLC-LLM~\cite{mlc-llm}, and llama.cpp~\cite{llama.cpp}, making it highly versatile.

\noindent\textbf{High extensibility.} The toolkit is highly modular and extensible, allowing easy adaptation~\footnote{All adaptations mentioned here have been implemented and results are shown in the appendix.} from integer quantization to floating-point quantization, from LLMs to VLMs~\cite{zhang2024visionlanguagemodelsvisiontasks}, from quantization to sparsification, and from dense models to Mixture-of-Expert (MoE) models~\cite{shazeer2017outrageouslylargeneuralnetworks}. This modularity ensures users can extend and customize the toolkit to meet their needs.

\noindent\textbf{Comprehensive evaluation.} LLMC enables comprehensive evaluation of quantized models, providing detailed performance metrics and analysis, \emph{e.g.}, PPL~\cite{alon2023detectinglanguagemodelattacks}, and data visualization analysis, \emph{e.g.,} Kurtosis value, quantization error, and outlier distribution. This thorough evaluation capability ensures that users can make informed decisions about the best quantization strategies for their models.

\section{Benchmarking LLM Quantization}
Powered by LLMC toolkit, we explore the quantization of LLMs from three distinct perspectives: the calibration data in~\autoref{sec:cali}, the algorithms in~\autoref{sec:algo}, and the data format of quantization in~\autoref{sec:format}. More explorations, \emph{e.g.}, extendability of LLMC, KV cache quantization, and inference speed can be found in the appendix.

\subsection{Experimental Settings}\label{sec:detail}
We first introduce experimental settings as follows. More implemental details with quantization preliminary can be found in the appendix.

\noindent\textbf{Models. } To demonstrate the generability of our benchmark, we access performance on LLaMA-2~\citep{touvron2023llama} and LLaMA-3~\citep{llama3modelcard} family, spanning model sizes from 7B to 70B for general language tasks. To broaden the scope of our evaluation, we show more results in the appendix, including ChatGLM~\citep{zeng2023glm-130b} for long context abilities, LLaVA-1.5 ~\cite{liu2023improved} for the multimodal task, Mixtral~\cite{jiang2024mixtral} as a representative of MoE models.

\noindent\textbf{Datasets. }We categorize the evaluation datasets into upstream and downstream datasets. For the upstream datasets, we employ WikiText2~\citep{wikidump} and C4~\citep{2019t5} dataset with the perplexity metric for evaluation, since perplexity can stably reflect the LLM's perfomance~\citep{dettmers2023case}. For the downstream tasks, we select examination tasks including MMLU~\citep{mmlu}, ARC-e~\citep{allenai:arc}, BoolQ~\citep{clark2019boolq}, HellaSwag~\citep{zellers2019hellaswag}, PIQA~\citep{piqa}, GPQA~\cite{rein2023gpqagraduatelevelgoogleproofqa}, MBPP~\citep{mbpp}, Human-Eval~\cite{chen2021codex}, the long context evaluation LongBench~\citep{bai2023longbench}, and multimodal evaluation MME~\cite{fu2023mme}. For the calibration data, to ensure a fair comparison, the vast majority of experiments use the same subset of the Pile~\cite{gao2020pile} validation set. We use the same calibration data number of 128 and the same sequence length of 512. We also find that different preprocessing methods of the calibration data can affect the quantization accuracy significantly. So, we use the same preprocessing method as in our open-source code.

\subsection{Impact of Calibration Data}\label{sec:cali}
With fair experimental settings, we first explore how calibration data impacts quantization accuracy. Prior studies~\cite{li2023norm, liu2023llmqat} highlight significant effects of different calibration datasets on quantized model performance. Yet, a systematic analysis of crucial factors is lacking. To address this, we identify and propose two key aspects to guide future calibration data selection.

\begin{figure}[ht!]
\vspace{-0.1in}
   \centering
       \includegraphics[width=1.\linewidth]{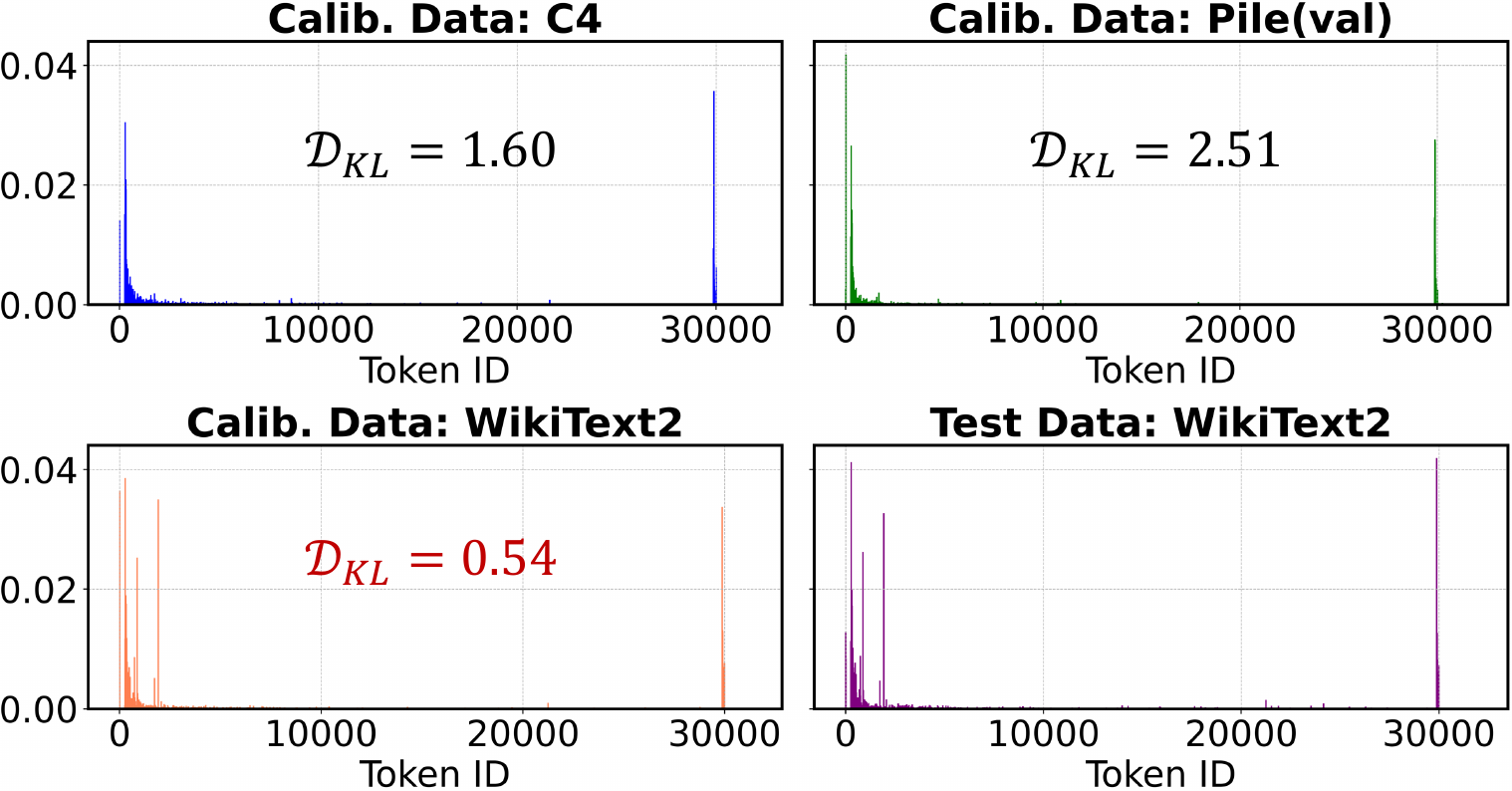}
        \vspace{-0.3in}
       \caption{Token distribution for calibration/test datasets. The y-axis shows frequency, the x-axis shows token ID, and ``$\mathcal{D}_{KL}$'' calculates the KL divergence between the calibration data and the specific test data: WikiText2.} %
   \label{fig:calib-data}
   \vspace{-0.1in}
\end{figure}

\noindent\textbf{Token distribution consistency.} Previous research~\citep {Cai2020ZeroQAN, zhang2021diversifying} focuses on synthesizing better distribution-matched calibration images to achieve higher performance for vision models. Derived from that view, we are the first to investigate the impact of the token distribution relationship between calibration and test data on model performance. As shown in \autoref{tab:cali-data} and \autoref{fig:calib-data}, we find that the performance of a model calibrated with data that more closely matches the token distribution of the test set tends to be superior. For instance, WikiText2 calibration data with 1.97 lower $D_{KL}$ achieves a $\approx$ 0.2 PPL decrease than Pile~(val) on the WikiText2 test data with GPTQ quantization. This finding indicates the importance of selecting calibration data with an aligned distribution for the data in practice.
\begin{table}[!ht]
    \renewcommand{\arraystretch}{1.2}
    \vspace{-0.1in}
    \centering
    \scalebox{0.6}{
    \begin{tabular}{c|l:l:l}
\toprule
\multicolumn{1}{c|}{\textbf{Calib. Data}} & \multicolumn{1}{c}{\textbf{GPTQ}} & \multicolumn{1}{c}{\textbf{AWQ}} & \multicolumn{1}{c}{\textbf{OmniQuant}} \\
\cmidrule(l){1-1}\cmidrule(l){2-2}\cmidrule(l){3-3}\cmidrule(l){4-4}
C4 &  6.323     &   6.173     &  5.717     \\
Pile (val) & 6.330  & 6.195 & 5.753 \\
WikiText2 & $\mathbf{6.133_{\mg{+0.568}}}$    & $\mathbf{6.144_{\mg{+0.156}}}$    & $\mathbf{5.697_{\mg{+0.516}}}$  \\

\bottomrule
\end{tabular}

    }
    \vspace{-0.1in}
    \caption{Impact of calibration data on performance across algorithms. We evaluate the PPL$\downarrow$ of WikiText2 test data, employing w3a16g128 GPTQ~\cite{frantar2022gptq} and AWQ~\cite{lin2023awq}, and w6a6 QmniQuant~\cite{shao2023omniquant} quantized LLaMA-2-7B. \mg{Data indices} show differences in results from randomly shuffling token order within each data entry.}
    \label{tab:cali-data}
    \vspace{-0.2in}
\end{table}

\begin{figure*}[ht!]
\vspace{-0.1in}
   \centering
       \includegraphics[width=1.\linewidth]{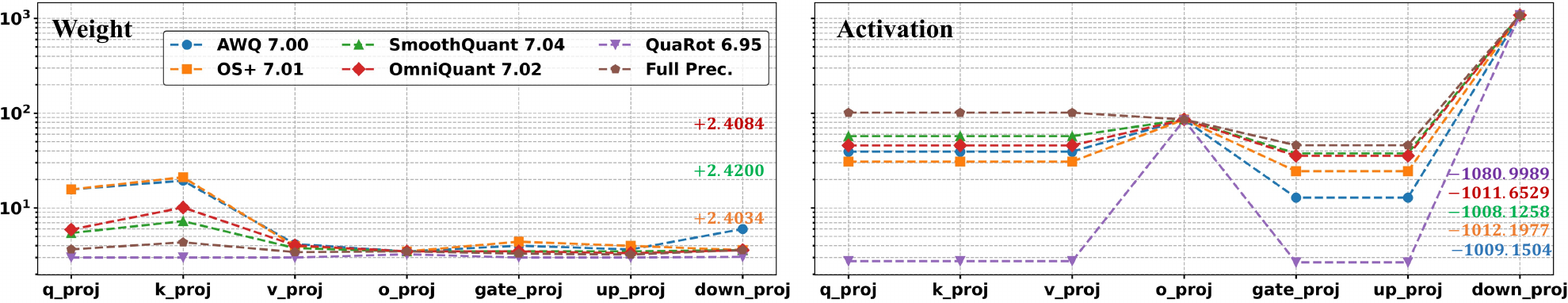}
        \vspace{-0.25in}
       \caption{Kurtosis value of weights~(Left) and input activations~(Right) with various layer types for different methods under w6a6 quantization. The legends denote the quantization method and its corresponding PPL on WikiText2. We do not employ transformation for \texttt{down\_proj} for a fair comparison, as only default AWQ and QuaRot include this position. The colorful values represent changes of $K$ after using transformation for \texttt{down\_proj} for all scaling-based methods, and online transformation for QuaRot. To be noted, we only mark numbers $>0.2$ for all the cases.} %
   \label{fig:k}
   \vspace{-0.2in}
\end{figure*}
\noindent\textbf{Intra-sentence logic.} Unlike vision models that utilize image calibration data, LLMs' calibration data consist of sequentially ordered token sequences that embody logical meaning. Therefore, we also conduct experiments to explore the impact of that logic on LLM quantization. Seeing from the \mg{data indices} in \autoref{tab:cali-data}, breaking the logic within the calibration data can cause a non-negligible accuracy drop. Notably, in this scenario, the robustness of learning/reconstruction-based algorithms such as GPTQ, and OmniQuant are lower than non-learning methods. Specifically, both exhibit $\times 3.3$ PPL increasing compared with AWQ. Overall, people should not seek or generate an illogical corpus to calibrate LLMs.

\subsection{Dive into the Quantization Algorithms}\label{sec:algo}
Besides calibration data, we could also methodically explore and benchmark LLM quantization algorithms equipped with our LLMC. Three main techniques for the field are outlier transformation, weight clipping, and weight reconstruction. However, how and how much they help under different scenarios remains unclear, as existing studies lack fair comparisons. Therefore, we will respectively discuss these methods in this section.

\subsubsection{How Does Transformation Influence Activation and Weight Outlier? }\label{sec:trans}
Most of the existing works aim to reduce the outliers via different kinds of equivalent transformation~\footnote{In this section, our experiments only employ transformation methods in each algorithm. We also apply transformation of AWQ to weight activation quantization.}, which can be categorized as scaling-based transformation, \emph{e.g.}, AWQ~\cite{lin2023awq}, SmoothQuant~\cite{xiao2023smoothquant}, OS+~\cite{wei2023outlier}, and OmniQuant~\cite{shao2023omniquant} and rotation-based transformation, for instance, QuaRot~\cite{ashkboos2024quarotoutlierfree4bitinference}. 

Scaling-based transformation typically involves searching for or learning a scaling vector to convert activation outliers into weights by optimizing the layer's quantization error. Conversely, the rotation-based transformation employs an Orthogonal matrix without accounting for output error. To thoroughly examine their effects, we analyze the kurtosis value~\footnote{Kurtosis value is defined as $K=\frac{1}{n}\sum^n_{i=1}(\frac{\boldsymbol{X}_i-\mu}{\sigma})^4$, where $\mu$ and $\sigma$ represent mean and variance of a tensor $\boldsymbol{X}$, to reflect outlier conditions~\cite{bondarenko2023quantizabletransformersremovingoutliers}.} of each layer after transformation, providing insights into their inherent mechanisms.

From \autoref{fig:k} and \autoref{tab:wo-k-ppl}, We observe three distinct findings. \textit{1) } Scaling-based transformation methods achieve lower $K$ for activations at the cost of higher $K$ for weights compared with full precision, which would induce a non-negligible performance degradation for lower-bit weight quantization, even with higher-bit activations can not eliminate the risk~(w6a6 > w4a8 in \autoref{tab:trans}). \textit{2) } $K$ for some specific positions like \texttt{down\_proj} layers is significantly higher than others. These positions have a pronounced impact on accuracy. For example, with \texttt{down\_proj} transformed~(evident lower $K$ in \autoref{fig:k}), salient improvements are gained as exhibited in \autoref{tab:trans}. \textit{3) } Although the rotation-based transformation reduces outliers by directly optimizing the tensor's outliers, it may not realize obvious accuracy improvement in some cases. From \autoref{tab:wo-k-ppl}, it is evident that the quantization error of output tensors is not minimized, as optimization did not focus on reducing output error, leading to a higher PPL.

\begin{table}[!ht]
\setlength{\tabcolsep}{0.5mm}
    \renewcommand{\arraystretch}{1.2}
    \vspace{-0.1in}
    \centering
    \scalebox{0.6}{
    \begin{tabular}{ccccccccc}
\toprule
\textbf{Method} & \texttt{q\_proj} & \texttt{k\_proj} & \texttt{v\_proj} & \texttt{o\_proj} & \texttt{gate\_proj} & \texttt{up\_proj} & \texttt{down\_proj} & PPL$\downarrow$\\
\midrule
Full Prec. &3.6505 &4.3354& 3.4174& 3.4720& 3.2991& 3.2300& 3.5845 & 6.14\\
\midrule
\multirow{2}{*}{AWQ} &4.9219& 6.1633& 3.4602& 3.4720& 3.3190& 3.2438& 4.3083& \multirow{2}{*}{8.57}\\
\cdashline{2-8}
&\cellcolor[gray]{0.92}0.9960& \cellcolor[gray]{0.92}0.9960& \cellcolor[gray]{0.92}0.9784& \cellcolor[gray]{0.92}0.9387& \cellcolor[gray]{0.92}0.9882& \cellcolor[gray]{0.92}0.9628& \cellcolor[gray]{0.92}0.9479 &\\
\midrule
\multirow{2}{*}{QuaRot} &2.9051 &2.9050 &2.9069 &2.9075 &2.9074 &2.9073 &2.9075 & \multirow{2}{*}{40.81} \\
\cdashline{2-8}
&\cellcolor[gray]{0.92}0.9962& \cellcolor[gray]{0.92}0.9967& \cellcolor[gray]{0.92}0.9797& \cellcolor[gray]{0.92}0.8286& \cellcolor[gray]{0.92}0.9764& \cellcolor[gray]{0.92}0.9579& \cellcolor[gray]{0.92}0.9230 &\\
\bottomrule
\end{tabular}

    }
    \vspace{-0.1in}
\caption{Comparison on $K$ and PPL on Wikitext2 of w3a16g128 LLaMA-3-8B for scaling-based transformation methods AWQ and rotation-based transformation method QuaRot. Due to the neglect of optimizing output quantization error~(cosine similarity in the \gr{gray cells}), QuoRot results in higher PPL even with fewer outlier issues.}
\label{tab:wo-k-ppl}
\end{table}
\begin{table}[!t]
\setlength{\tabcolsep}{1mm}
    \renewcommand{\arraystretch}{1.1}
    \vspace{-0.1in}
    \centering
    \setlength{\tabcolsep}{1.5mm}
    \resizebox{ \linewidth}{!}{
    \begin{tabular}{cccccccccc}
\toprule
 \multicolumn{2}{c}{AWQ} & \multicolumn{2}{c}{SmoothQuant} & \multicolumn{2}{c}{OS+} & \multicolumn{2}{c}{OmniQuant} & \multicolumn{2}{c}{QuaRot} \\
\cmidrule(l){1-2}\cmidrule(l){3-4}\cmidrule(l){5-6}\cmidrule(l){7-8}\cmidrule(l){9-10}
w4a8 & w6a6 & w4a8 & w6a6 & w4a8 & w6a6 & w4a8 & w6a6 & w4a8 & w6a6 \\\midrule
8.60 & 7.00 & 8.85 & 7.04 & 8.55 & 7.01 & 8.83 & 7.02 & 9.77 & 6.95 \\\cdashline{1-10}
\rowcolor[gray]{0.92}7.77 & 6.79 & 7.92 & 6.85 & 7.76 & 6.81 & 7.92 & 6.83 & 9.43 & 6.74 \\
\bottomrule
\end{tabular}

    }
    \vspace{-0.1in}
\caption{PPL on Wikitext2 for different transformation methods with or without transforming \texttt{down\_proj} layers for LLaMA-3-8B. The \gr{gray raw} indicates the results are obtained with \texttt{down\_proj} layers transformed.}
\label{tab:trans}
 \vspace{-0.18in}
\end{table}

\subsubsection{When Should We Utilize the Weight Clipping? }
The technique of weight clipping, restricting the range of weight values before quantization, has been recognized for its contribution to maintaining better performance~\cite{lin2023awq, du2024bitdistiller, shao2023omniquant} for the quantization process. Here, we analyze its application situations under two different scenarios.

\begin{figure}[ht!]
\vspace{-0.1in}
   \centering
       \includegraphics[width=1.\linewidth]{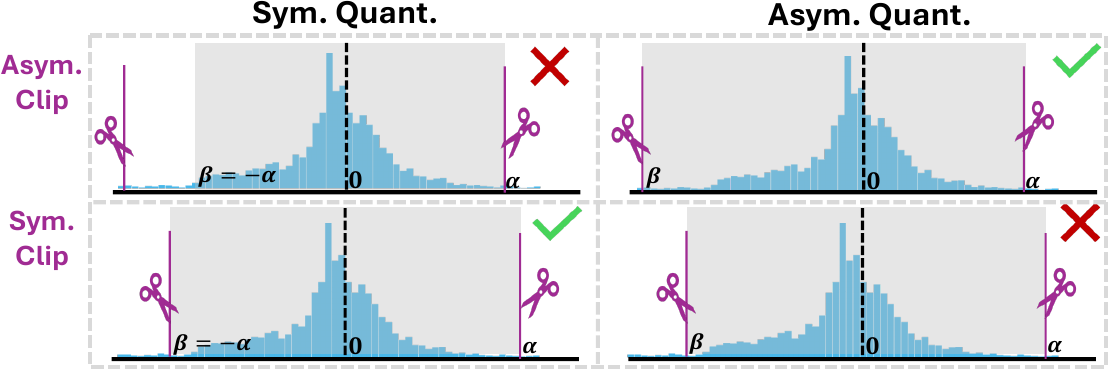}
        \vspace{-0.2in}
       \caption{Comparison between asymmetric and symmetric \pp{weight clipping} \emph{w.r.t.} asymmetric/symmetric quantization. After weight clipping, we obtain the final range of tensor to quantize as depicted in the \gr{solid gray box} related to asymmetric/symmetric quantization.} %
   \label{fig:align-clip}
\end{figure}

\noindent\textbf{Symmetric or asymmetric.} Clipping and quantization can be divided into symmetric or asymmetric categories. However, previous studies~\cite{lin2023awq, liu2024qllm} always neglect their relationships and employ wrong patterns. As shown in \autoref{fig:align-clip}, we can observe that symmetric clipping with symmetric quantization maintains more information~(\emph{i.e.}, \gr{solid gray box}) than with asymmetric quantization, and for asymmetric clipping vice versa. This finding can help improve current methods with significant accuracy recovery, especially for extremely lower bit-width. For instance, in \autoref{tab:awq-clip}, default AWQ, applying asymmetric quantization with symmetric clipping, results in a 6.8e4 PPL score and performance~\footnote{Without special claims, we calculate average accuracy on five downstream tasks: MMLU, ARC-e, BoolQ, HellaSwag, and PIQA, and average PPL on WikiText2 and C4 in the paper. Detailed data is presented in the appendix \autoref{detail}.} declines of 48.11\% for 2-bit LLaMA-2-70B compared with 3-bit configuration. Conversely, equipping with asymmetric clipping, AWQ in LLMC achieves 42.47\% accuracy upswings with admissible PPL.
\begin{table}[!t]
    \renewcommand{\arraystretch}{1.3}
    \setlength{\tabcolsep}{1mm}
    \vspace{-0.1in}
    \centering
    \resizebox{ \linewidth}{!}{
    \begin{tabular}{clcccc}
\toprule

\multirow{2}{*}{\textbf{\#Bits}} & \multirow{2}{*}{\textbf{Method}} & \multicolumn{2}{c}{LLaMA-2-7B} & \multicolumn{2}{c}{LLaMA-2-70B}\\\cmidrule(l){3-4}\cmidrule(l){5-6}
& & Avg. PPL$\downarrow$ & Avg. Acc.$\uparrow$ & Avg. PPL$\downarrow$ & Avg. Acc.$\uparrow$ \\
\midrule
\multirow{2}{*}{w3a16g128}& AWQ & 7.25& 61.18 & 4.90& 80.95 \\
\cdashline{2-6}
& AWQ \small{\emph{w/} asym. clip} & 7.21 & 61.59 &4.89 & 81.07\\\cmidrule(l){1-6}
\multirow{2}{*}{w2a16g64} & AWQ & 1.8e5 & 37.69 & 6.8e4 & 32.84  \\
\cdashline{2-6}
& AWQ \small{\emph{w/} asym. clip} &13.26 & 48.77 & 6.49 & 75.31 \\

\bottomrule
\end{tabular}

    }
    \vspace{-0.1in}
    \caption{Impact of asymmetric/symmetric weight clipping. We evaluate the average accuracy and the average PPL here. ``asym. clip'' means we employ asymmetric clipping.}
    \label{tab:awq-clip}
    \vspace{-0.25in}
\end{table}

\noindent\textbf{Bit-width.} Besides different combinations of quantization and clipping, we also investigate the impact of clipping with different bit-width. 
\begin{table}[t]
\renewcommand{\arraystretch}{1.1}
    \vspace{-0.1in}
    \centering
    \setlength{\tabcolsep}{1mm}
    \resizebox{ \linewidth}{!}{
    
\begin{tabular}{ccccccccc}
\toprule
\multirow{2}{*}{\textbf{Model}} & \multicolumn{2}{c}{w3a16g128} & \multicolumn{2}{c}{w4a16g128}  & \multicolumn{2}{c}{w6a6} & \multicolumn{2}{c}{w8a8}\\
 \cmidrule(lr){2-3} \cmidrule(lr){4-5} \cmidrule(lr){6-7} \cmidrule(lr){8-9}
&  \emph{w/} clip & \emph{w/o} clip & \emph{w/} clip &\emph{w/o} clip  & \emph{w/} clip & \emph{w/o} clip & \emph{w/} clip & \emph{w/o} clip\\
\midrule
\rowcolor[gray]{0.92}\cellcolor{white}\multirow{2}{*}{LLaMA-3-8B} & 11.74 & 11.23 & 11.99 & 17.42 & 10.35 & 9.46 & 10.73 & 10.35 \\\cdashline{2-9}
& 30.60 & 24.80 & 40.60 & 42.20 & 40.60 & 39.40 & 43.80 & 43.80 \\\midrule
\rowcolor[gray]{0.92}\cellcolor{white}\multirow{2}{*}{LLaMA-3-70B} & 8.08 & 7.57 & 9.09 & 11.62 & 26.38 & 25.75 & 16.79 & 16.66\\\cdashline{2-9}
& 54.00 & 54.20 & 59.20 & 60.00 & 58.20 & 58.20 & 60.20 & 57.60\\
\bottomrule
\end{tabular}

    }
    \vspace{-0.1in}
\caption{Impact of weight clipping under various bit-width. We employ AWQ for weight-only and OS+ for weight activation quantization with or without clipping as methods here. Accuracy on GPQA is highlighted in \gr{gray rows}, and the rest for MBPP.}
\label{tab:clip-bit-main}
\vspace{-0.1in}
\end{table}
From \autoref{tab:clip-bit-main}, weight clipping does not show superiority across all bit-widths. \textit{1) For higher bit~(4-bit) weight-only quantization, clipping has a side-effect, unlike improvement for lower-bit~(3-bit).} We hypothesize that in 4-bit quantization, weight clipping causes more information loss than quantization rounding. However, for 3-bit quantization, quantization rounding has a greater impact. \textit{2) For weight activation quantization, suitable clipping exhibits positive effects whatever bit-width.} We ascribe this for clipping anomalous values effectively adjusting the majority of weights~(\emph{i.e.}, moderate and small elements). Accounting for hard-quantized and considerably influential activations, this approach significantly reduces the output errors resulting from the multiplication of quantized large activations with well-adjusted weights~\footnote{Activation outliers make huge performance deterioration can be found in \texttt{LLM.int8()}~\cite{dettmers2022llm}.}, which greatly reduce the impact of these quantized activations.

\subsubsection{Should We Combine Transformation and Reconstruction?}
Apart from transformation and clipping, the reconstruction-based method like GPTQ~\cite{frantar2022gptq} is also widely used to quantize weights. This method iteratively updates the unquantized weights to compensate for the impact of the current quantized weights, thereby minimizing the output quantization error. Some recent transformation methods~\cite{ashkboos2024quarotoutlierfree4bitinference, lin2023awq} integrate this technique to demonstrate their extendability.

Nevertheless, we find that a significant and obvious accuracy from this combination is not usually the case. From \autoref{tab:recon-trans}~\footnote{Clipping for AWQ here is canceled to expel distractions.}, we conclude that: \emph{1)} the scaling-based transformation like AWQ \emph{w/} GPTQ shows moderate improvement for LLaMA-3-8B. \emph{2)} However, The rotation-based method QuaRot \emph{w/} GPTQ far surpasses QuaRot alone, even with 28.94\% accuracy boost for 3-bit LLaMA-3-8B.
\begin{table}[t!]
    \renewcommand{\arraystretch}{1.2}
    \centering
    \scalebox{0.65}{

\begin{tabular}{lccccc}
\toprule
 \textbf{Metric} & GPTQ & AWQ & AWQ \small{\emph{w/} GPTQ} & QuaRot & QuaRot \small{\emph{w/} GPTQ} 
\\\midrule
Avg. PPL$\downarrow$& 10.67 & 10.98 & 10.55 & 50.00 & 10.35
\\\cdashline{1-6}
Avg. Acc.$\uparrow$ & 71.96 & 70.72 &72.72 & 45.90 & 74.84
\\
\bottomrule

\end{tabular}

    }
    \vspace{-0.1in}
\caption{Impact of reconstruction~(GPTQ) combined with scaling~(AWQ) and rotation-based~(QuaRot) transformations for w3a16g128 LLaMA-3-8B.}
\label{tab:recon-trans}
 \vspace{-0.05in}
\end{table}
\begin{figure}[t!]
   \centering       \includegraphics[width=\linewidth]{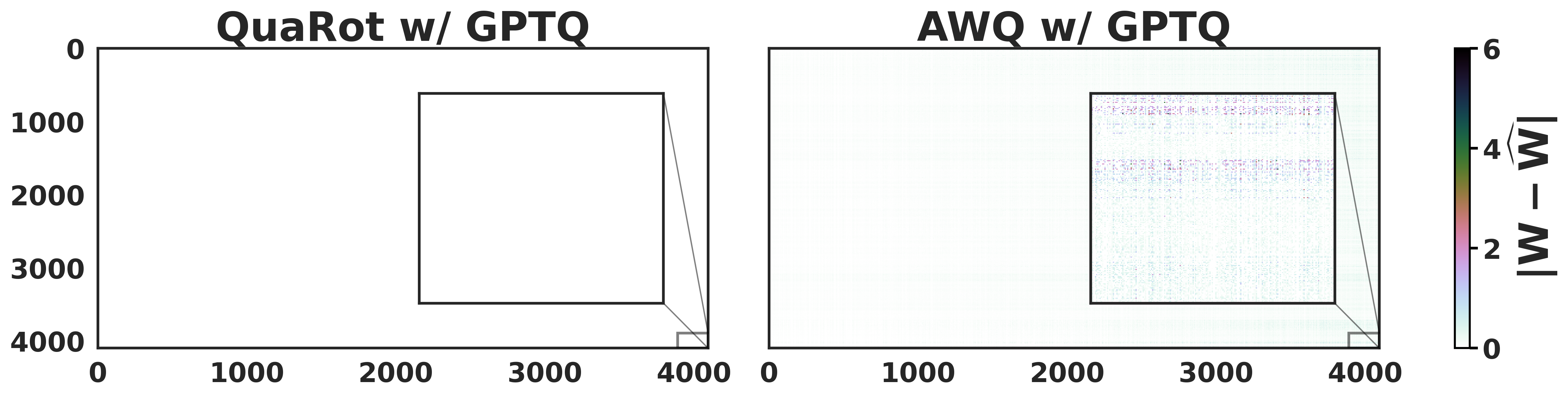}
        \vspace{-0.3in}
       \caption{Visualization of relative quantization errors for the weight of \texttt{q\_proj} in the first block for w3a16g128 LLaMA-3-8B. $\widehat{\boldsymbol{W}}$ represents the quantized counterpart of the weight $\boldsymbol{W}$.} %
   \label{fig:quant-error}
   \vspace{-0.2in}
\end{figure}
The inherent reason might lie in two aspects: \textit{1)} Scaling-based transformation methods may amplify weight outliers~\footnote{$K$ analysis in \autoref{sec:trans} verifies this.}. This gives rise to a larger challenge for iterative compensation during the reconstruction, especially weights in rear columns which GPTQ can not properly deal with~\footnote{This can be found in QUIK~\citep{ashkboos2023quik}}. However, QuaRot, which effectively eliminates weight outliers, pairs well with GPTQ. From \autoref{fig:quant-error}, the steeper quantization error of later weight columns for AWQ \emph{w/} GPTQ compared with QuaRot \emph{w/} GPTQ validates our analysis.  \textit{2)} Rotation-based transformation only aims to decrease tensor outliers without considering output errors, so the kurtosis value is significantly reduced. However, for weight-only quantization, outliers in the activation might amplify the error in quantized weights~\footnote{The importance of salient activation is described in AWQ.}, leading to obvious output discrepancy. GPTQ exactly considers the output error through approximated Hessian matrix, and thus can always complement rotation-based transformation. As in \autoref{tab:quarot-ana}, QuaRot \emph{w/} GPTQ performing a much higher cosine similarity between the output of the corresponding layer and its quantized counterpart helps confirm our analysis.
\begin{table}[tp!]
\setlength{\tabcolsep}{1mm}
\renewcommand{\arraystretch}{1.3}
    \centering
    \scalebox{0.56}{
    
\begin{tabular}{cccccccc}
\toprule
\multirow{1}{*}{\textbf{Method}} & \texttt{q\_proj} & \texttt{k\_proj} & \texttt{v\_proj} & \texttt{o\_proj} & \texttt{gate\_proj} & \texttt{up\_proj} & \texttt{down\_proj} \\
\midrule
QuaRot & 0.9962 & 0.9967 & 0.9797 & 0.8286 & 0.9764 & 0.9579 & 0.9230 \\
\midrule
\multirow{1}{*}{QuaRot \small{\emph{w/} GPTQ}} & 0.9971 & 0.9975 & 0.9847 & 0.9476 & 0.9895 & 0.9791 &0.9529 \\
\bottomrule
\end{tabular}

    }
    \vspace{-0.1in}
\caption{
Fine-grained analysis comparing QuaRot and QuaRot \emph{w/} GPTQ in w3a16g128 LLaMA-3-8B. We report the output cosine similarity between the original layer and the quantized layer.
}
\label{tab:quarot-ana}
\vspace{-0.1in}
\end{table}

\subsection{Integer or Floating-point Quantization?}\label{sec:format}
The above-mentioned algorithms are based on integer~(INT) quantization. Although traditional INT quantization has received widespread adoption in the industry, floating-point~(FP) quantization has emerged as a rising alternative. This is attributed to its superior accuracy and high flexibility, offering advantages for handling long-tailed distributions.
\begin{table}[t!]
    \setlength{\tabcolsep}{1mm}
    \renewcommand{\arraystretch}{1.3}
    \centering
    \scalebox{0.53}{

\begin{tabular}{ccccccccccc}
\toprule
     \multirow{2}{*}{\makecell{Full\\Prec.}} & \multicolumn{2}{c}{w3a16g128} & \multicolumn{2}{c}{w4a16g128} & \multicolumn{2}{c}{w4a16} & \multicolumn{2}{c}{w4a4} & \multicolumn{2}{c}{w6a6}\\\cmidrule(l){2-3}\cmidrule(l){4-5}\cmidrule(l){6-7}\cmidrule(l){8-9}\cmidrule(l){10-11}
     & Naive & AWQ & Naive & AWQ & Naive & AWQ & Naive & SmoothQuant & Naive & SmoothQuant\\\midrule
     \rowcolor[gray]{0.92}\cellcolor{white}\multirow{2}{*}{5.47} & 6.66 & 6.19 & 5.78 & 5.59 & 6.11 & 5.81 & NaN & NaN & 6.86 & 6.77 \\\cdashline{2-11}
     & 6.89 & 6.38 & 5.70 & 5.63 & 5.89 & 5.75 & 90.85 & 16.35 & 5.56 & 5.56 \\
     
     \bottomrule
\end{tabular}
}
    \vspace{-0.1in}
    \caption{PPL for LLaMA-2-7B weight-only quantization and weight-activation INT~(\gr{gray rows})/FP~(whight rows) quantization on WikiText2. Naive means simple round-to-nearest quantization.}
    \label{tab:fp-quantization}
    \vspace{-0.2in}
\end{table}

\autoref{tab:fp-quantization} reports the detailed FP quantization results for LLMs. For the weight-activation quantization, FP quantization consistently surpasses INT quantization by a large margin as it can better overcome the outlier issue. It is worth noting that under w4a4, the INT quantization suffers from non-trivial performance degradation while FP quantization improves to a usable level. Conversely, when applying weight-only quantization, the FP quantization achieves worse performance under ultra-low-bit ($\le$ 3-bit) or small group size. These findings indicate that: \textit{1)} the positive zero and negative zero in FP format constrain the representation capability of this quantization type, particularly under low-bit. \textit{2)} the range of small group size is more uniform, which is unsuitable for FP quantization. \textit{3)} the symmetric FP quantization struggles to deal with the asymmetry in LLMs.

\section{Additional Results and Discussions}
\noindent\textbf{Impact of quantization for fine-tuning.} We conduct experiments for quantization on LLaMA-3-8B with supervised fine-tuning (SFT) on Evol-instruction-66k~\footnote{\url{https://huggingface.co/datasets/codefuse-ai/Evol-instruction-66k}} to analyze the impact. We choose ms-swift~\cite{zhao2024swiftascalablelightweightinfrastructure} as the finetuning framework. Additionally, we set the learning rate to 2e-6 with a mini-batch size of 2 and trained the model for 1 epoch on 16 40G A800 GPUs. After fine-tuning, we employ w4a16 naive quantization and AWQ to quantize the model. We choose HumanEval~\cite{chen2021evaluating} and HumanEval-X~\cite{zheng2023codegeex} for evaluation. As illustrated in \autoref{tab:fine-tuning}, quantization leads to more severe accuracy drops for the SFT model than the base model. This might be caused by the limited fine-tuning data and more in-depth analyses are needed in the future. Moreover, an advanced algorithm, \emph{i.e.}, AWQ brings obvious improvements compared to Naive quantization for the SFT model.
\begin{table}[h!]
\centering
\vspace{-0.1in}
\setlength{\tabcolsep}{2mm}
\resizebox{ \linewidth}{!}
{
\begin{tabular}{lccc}
\toprule
Test Data & Base/SFT & Base/SFT+Naive & Base/SFT+AWQ\\
\midrule
HumanEval& 23.78/49.39&	19.51/42.07	&21.34/46.34\\
HumanEval-X &32.81/41.58	&26.47/36.10&	26.83/39.27\\
\bottomrule
\end{tabular}
}
\vspace{-0.1in}
    \caption{Accuracy of Base/SFT models after quantization. ``Base'' denotes LLaMA-3-8B. We report the average accuracy of 5 languages in HumanEval-X.}
    \label{tab:fine-tuning}
\vspace{-0.1in}
\end{table}

\noindent\textbf{Impact of calibration data for VLMs.} Besides LLMs, we further present the impact of calibration data for LLaVA-7B~\cite{liu2023improved} here. The results in \autoref{tab:calib-data-vlm} indicate that we should collect text and vision data together for VLM quantization.
\begin{table}[h!]
\centering
\vspace{-0.1in}
\setlength{\tabcolsep}{2mm}
\resizebox{0.7\linewidth}{!}
{
\begin{tabular}{lcc}
\toprule
Method & Perception &	Cognition \\
\midrule
FP	& 1477.60	& 283.21 \\ 
\hdashline
Calib. Data: Pile (val) &	1437.94 & 	274.64 \\
Calib. Data: T\&V  &	1470.93	& 286.78 \\
\bottomrule
\end{tabular}
}
\vspace{-0.1in}
    \caption{Impact of calibration data for VLMs. We employ w4a16 AWQ. ``T\&V'' denotes MS-COCO~\cite{lin2014microsoft} and TextVQA~\cite{singh2019towards}.}
    \label{tab:calib-data-vlm}
\vspace{-0.15in}
\end{table}

\noindent\textbf{Accuracy alignment with the existing methods.} Except for the PPL alignment results in \autoref{sec:align}, we further conduct downstream experiments for LLaMA-2-7B to prove our reproducibility (experimental details in the appendix). As illustrated in \autoref{tab:alig-wo-down} and \autoref{tab:alig-wa-down}, our LLMC is reliable in reproducing the outcomes of existing quantization methods.
\begin{table}[h!]
\centering
\vspace{-0.1in}
\setlength{\tabcolsep}{2mm}
\resizebox{0.7\linewidth}{!}
{
\begin{tabular}{lcccc}
\toprule
w4a16g128& MMLU	&BoolQ	&ARC-e&	PIQA\\
\midrule
AWQ	&46.36&	71.25&	54.14&	77.04\\
AWQ-LLMC&	46.47&	71.62	&53.96	&77.26 \\
\hdashline
GPTQ&	43.36	&72.81	&51.50	&77.86 \\
GPTQ-LLMC	&43.40	&72.91	&51.50	&77.75 \\
\bottomrule
\end{tabular}
}
\vspace{-0.1in}
    \caption{Alignment for weight-only quantization. ``-LLMC'' represents the results are reproduced with our toolkit LLMC.}
    \label{tab:alig-wo-down}
\vspace{-0.1in}
\end{table}
\begin{table}[h!]
\centering
\setlength{\tabcolsep}{1.5mm}
\resizebox{\linewidth}{!}
{
\begin{tabular}{lcccc}
\toprule
w8a8&	MMLU&	BoolQ	&ARC-e&	PIQA\\
\midrule
SmoothQuant	&46.17	&69.76	&49.03	&77.26 \\
SmoothQuant-LLMC	&46.28&	69.08	&50.97	&77.26\\
\hdashline
QuaRot \small{\emph{w/} GPTQ}&	46.38&	71.50	&52.73	&77.75\\
QuaRot-LLMC + \small{\emph{w/} GPTQ-LLMC.}	&46.42&	70.61&	53.26&	77.97\\

\bottomrule
\end{tabular}
}
\vspace{-0.1in}
    \caption{Alignment for weight-activation quantization.}
    \label{tab:alig-wa-down}
\vspace{-0.2in}
\end{table}

\noindent\textbf{Role of model scales.} Besides LLaMA-2 and LLaMA-3 families, we also conduct experiments for quantizing different LLM families, \emph{e.g.}, SmolLM-135M/350M/1.7B~\footnote{\url{https://huggingface.co/blog/smollm}}, MiniCPM-1B/2B~\cite{hu2024minicpm}, and Qwen-2-0.5B/1.5B~\cite{yang2024qwen2} in \autoref{sec:scales}. We find that low-bit quantization causes more performance degradation for homology models with a larger size. This phenomenon is counter-intuitive and needs to be further explored. Besides, higher precision quantization, \emph{e.g.}, w8a8 or w4a16 leads to subtle accuracy drops for LLMs across all sizes. We will explore the role of scale for larger LLMs in the future.

\noindent\textbf{Pipeline of LLMC.} Basically, our LLMC receives an FP LLM and calculates its quantization parameters with advanced algorithms. Finally, this tool can export the model with quantization parameters to the quantization format compatible with a specific backend like vLLM~\cite{kwon2023efficient}. The detailed usage can be found in the official document~\footnote{\url{https://llmc-en.readthedocs.io/en/latest/}}. Additionally, LLMC can provide quantization analyses and PPL evaluations for those quantized LLMs. With this tool, people can produce various compressed industrial models deployed on different hardware~\footnote{Inference efficiency of compressed models can be found in \autoref{appendix:inference_speed}.}.

\section{Conclusion}
This paper introduces LLMC, a user-friendly and versatile toolkit for LLM compression. Supported by the toolkit, a series of observations and analyses were conducted, providing valuable and novel insights and suggestions for the community.

\section*{Acknowledgements}
We sincerely thank the anonymous reviewers for their serious reviews and valuable suggestions. This work was supported by the Beijing Municipal Science and Technology Project (No. Z231100010323002).

\bibliography{custom}
\clearpage
\appendix
\renewcommand{\contentsname}{\makecell{Content}}
\onecolumn
\tableofcontents 
\section{Appendix} 
\begin{table*}[!th]
    \renewcommand{\arraystretch}{1.8}
    \centering
    \scalebox{0.65}{\begin{tabular}{lllccc}
\toprule
\textbf{Technique} & \textbf{Approach} & \textbf{Strategy} & \textbf{Eq. Trans.}  & \textbf{Algorithm} \\
\midrule
\multirow{5}{*}{\textbf{\textsc{Transformation}}} & \multirow{2}{*}{\og{\textit{Rule-based}}} & $\vs=\mathrm{max}(|\mX|^{\gamma})/\mathrm{max}(|\mW|^{1-\gamma}), \gamma=0.5, 0.75, ...$ & \ding{51}   & SmoothQuant\cite{xiao2023smoothquant}  \\
& & $\boldsymbol{Q}$, where $\boldsymbol{Q}\boldsymbol{Q}^T=\boldsymbol{I}$ and $|\boldsymbol{Q}|=1$ & \ding{51} & QuaRot~\cite{ashkboos2024quarotoutlierfree4bitinference}\\
\cdashline{2-6}
   & \multirow{2}{*}{\red{\textit{Search-based}}} & $\vs=\mathrm{max}(|\mX|^{\gamma})/\mathrm{max}(|\mW|^{1-\gamma})$, grid search for $\gamma \in [0, 1]$  & \ding{51}    & AWQ\cite{lin2023awq}  \\
   & & $\vs=\mathrm{max}(1.0, \mathrm{max}(\mX) / t)$, grid search for $t$  & \ding{51}   &  OS+\cite{wei2023outlier}  \\
   \cdashline{2-6}
   & \mg{\textit{Learnining-based}} & $\vs=\argmin_{\vs} \mathcal{L}, \vs \leftarrow \vs - \eta \frac{\partial \mathcal{L}(\vs)}{\partial \vs}$ & \ding{51}    & OmniQuant\cite{shao2023omniquant} \\
\midrule 
\multirow{3}{*}{\textbf{\textsc{Clipping}}} & \og{\textit{Rule-based}} & $\alpha=1, \beta=1$ & \ding{51}  &    \makecell{SmoothQuant\cite{xiao2023smoothquant},\\ OS+\cite{wei2023outlier}, \\GPTQ\cite{frantar2022gptq}, \\QuaRot~\cite{ashkboos2024quarotoutlierfree4bitinference}} \\
\cdashline{2-6}
   & {\red{\textit{Search-based}}} & grid search for $\alpha=\beta \in[0, 1]$ & \ding{55}    & AWQ\cite{lin2023awq} \\
   \cdashline{2-6}
   & \mg{\textit{Learning-based}} & $\alpha, \beta=\argmin_{\alpha, \beta} \mathcal{L}, \alpha \leftarrow \alpha - \eta \frac{\partial \mathcal{L}(\alpha)}{\partial \alpha},  \beta \leftarrow \beta - \eta \frac{\partial \mathcal{L}(\beta)}{\partial \beta}$ & \ding{55}   & OmniQuant\cite{shao2023omniquant} \\
\midrule
\multirow{1}{*}{\textbf{\textsc{Reconstruction}}} & \gr{\textit{Hessian-based}} & $\mW \leftarrow \mW - \mE\mH^{-1}, \mH^{-1} = \left(2\mX{\mX}^\top + \lambda \mI\right)^{-1}$ & \ding{55}    & GPTQ\cite{frantar2022gptq}  \\
\bottomrule
\end{tabular}

}
    \vspace{-0.1in}
    \caption{Detailed comparison of the three main strategies in the main text. \textbf{Eq. Trans.} indicates whether the algorithm is an equivalent transformation. $\gamma$ is the scaling factor. $\vs$ and $\boldsymbol{Q}$ represent transformation vector and matrix. $\boldsymbol{I}$ is the identity matrix. $\mathcal{L}$ is the loss function with the learning rate $\eta$. $\alpha$ and $\beta$ mean clipping minimum and maximum value. $\boldsymbol{H}$ is Hessian matrix, and $\boldsymbol{E}$ denotes quantization errors calculated with $\boldsymbol{H}$. $\lambda$ is the decay coefficient.}
    \label{tab:comparison-algorithm}
\end{table*}
\twocolumn 
\subsection{Preliminary for Quantization}
A complete uniform quantization process can be formulated by:
\begin{equation}
\begin{split}
    \bar{\bm{w}} &= \mathrm{clip}(\left\lfloor \frac{\bm{w}}{s} \right\rceil+z,N_{min}, N_{max}), \\
    \hat{\bm{w}} &= s\cdot (\bar{\bm{w}}-z),
\end{split}
\label{eq:naive quant}
\end{equation}
where $s\in\mathbb{R}_+$ and $z\in\mathbb{Z}$ are called \textit{scale} and \textit{zero-point}, respectively. $\lfloor\cdot\rceil$ rounds the continuous numbers to the nearest integers. Eq.~\ref{eq:naive quant} first quantizes the weights or activations into the target integer range $[N_{\min}, N_{\max}]$ and then de-quantizes the integers to the original range. 

Naive quantization can be split into four dimensions: bit-width, symmetric/asymmetric, group size, and dynamic/static.

\noindent\textbf{Bit-width}: Given $t$ bits, $[N_{\min}, N_{\max}]$ is determined by $[-2^{t-1}, 2^{t-1}-1]$. In this paper, the notion ``w$x$a$y$'' is employed to represent the bit-widths of weights ``w'' and activations ``a'';

\noindent\textbf{Symmetric or asymmetric.} For asymmetric quantization, a zero-point value $z$ will usually be introduced to represent the floating-point zero. Otherwise, the symmetric quantization does not have that adjustable $z$ to adapt various ranges; 

\noindent\textbf{Group size.} \citet{shen2020q} first proposes group-wise quantization, which divides each channel of a weight~\footnote{We denote weight $\mW\in \mathbb{R}^{out\times in}$. The first/second dimension of $\mW$ represents output/input channels. Notably, we ignore the batch size dimension for activation $\mX\in\mathbb{R}^{n\times d}$, where $n$ means token number, $d$ means hidden size.} into different groups and employs a different set of scale and zero-point for each group $\mW_{i,j:j+g}$ with group size $g$. However, per-tensor~($\mW_{:,:}$) quantization or per-channel~($\mW_{i,:}$) quantization can be also seen as group-wise quantization with a larger group size; 

\noindent\textbf{Dynamic or static.} Due to variance in activation range for LLM, \citet{yao2022zeroquant} first introduces token-wise~($\mX_{i,:}$) quantization for activation, which dynamically calculates the min/max range for each token during model inference. We also measure dynamic/static per-tensor activation quantization to make a comprehensive comparison. 

As outlined in \autoref{tab:comparison-algorithm}, we also summarize the three strategies, \emph{e.g.}, transformation, clipping, and reconstruction in the main text and define their behavior. Additionally, for the equivalence transformation categories OS+ and OmniQuant, considering that we are using the LLaMA series models (which have layers without bias), we aim to avoid introducing additional computations into the model's inference process. Therefore, we have decided not to explore the shift operation involved in these two methods. 

\subsection{More Implementation Details}\label{sec:more-detail}
Unless otherwise specified, our implementation adopts asymmetric quantization for both activations and weights. Specifically, we apply per-token dynamic quantization for activations and static quantization for weights. g128 and g64 represent two commonly used settings in group weight quantization, indicating group sizes of 128 and 64, respectively. In line with previous works~\citet{shao2023omniquant,liu2024qllm,ashkboos2024quarotoutlierfree4bitinference}, For OmniQuant, the learning rate for weight clipping and transformation is $5e^{-3}$ and $1e^{-2}$ during the reconstruction phase. We follow the default setting of 20 learning epochs. Besides, we employ the evaluation tool OpenCompass~\citep{2023opencompass} with LightLLM~\citep{lightllm} as the backend on Nvidia A100 80G GPU to benchmark downstream tasks. Additionally, we evaluate PPL with 2048 sequence length in our own LLMC.

\subsection{PPL Alignment with the Existing Methods}\label{sec:align}
\begin{table}[ht!]
\centering
\setlength{\tabcolsep}{3mm}
\resizebox{ \linewidth}{!}
{
\begin{tabular}{ccccc}
\toprule
Method & Calib. Data & Sequence Length & Number of Samples & Seed \\
\midrule
GPTQ & C4  & 2048 & 128 & 0 \\
AWQ & Pile (val)  & 512 & 128 & 42 \\
Omniquant & Wikitext2  & 2048 & 128 & 2 \\
Smoothquant & Pile(val)  & 512 & 128 & 42 \\
OS+ & Pile (val) & 512 & 128 & 42 \\
Quarot & Wikitext2  & 2048 & 128 & 0 \\
Wanda & pileval & 512 & 512 & 42 \\ 
\bottomrule
\end{tabular}}
\vspace{-0.1in}
\caption{Calibration and hyperparameter settings in our alignment experiments.}
\label{settings}
\end{table}
In this section, we conduct some alignment experiments with several established quantization algorithms~(LLMC \emph{vs.} original paper/codes). Our experimental settings are the same as the original paper or default settings of their open-source codes~(as shown in \autoref{settings}). These experimental results are summarized in \autoref{w_only}, \autoref{w_a_1}, \autoref{w_a_2}, and \autoref{w_a_3}. The performance from the tables illustrates that our LLMC tool achieves performance almost identical to the original quantization algorithms reported in the literature. By employing these experiments, we demonstrate that our tool is not only effective but also reliable in reproducing the outcomes of existing quantization methods. This ensures that our contributions are both credible and valuable to the ongoing research in LLM quantization.
\begin{table}[th!]
\centering
\setlength{\tabcolsep}{5mm}
\resizebox{ \linewidth}{!}
{
\begin{tabular}{lccc}
\toprule
Method & w4g128 & w3g128 & w2g64 \\
\midrule
GPTQ & 5.62 & 6.32 & 14.97 \\
GPTQ-LLMC & 5.62 & 6.32 & 14.97 \\
\cdashline{1-4}
AWQ & 5.60 & 6.24 & 2.16e5 \\
AWQ-LLMC & 5.60 & 6.24 & 2.16e5 \\
\cdashline{1-4}
OmniQuant & 5.59 & 6.09 & 9.53 \\
OmniQuant-LLMC & 5.59 & 6.09 & 9.53 \\
\bottomrule
\end{tabular}}
\vspace{-0.1in}
\caption{Wikitext2 PPL alignment results of weight-only asymmetric quantization of LLaMA-2-7B Model. ``-LLMC'' means our implementation with the LLMC toolkit.}
\label{w_only}
\end{table}
\begin{table}[th!]
\centering
\setlength{\tabcolsep}{6mm}
\resizebox{ \linewidth}{!}
{
\begin{tabular}{lccc}
\toprule
Method & w8a8 & w6a6 & w4a4 \\
\midrule
OmniQuant & 5.49 & 5.70 & 12.21 \\
OmniQuant-LLMC & 5.49 & 5.70 & 12.23 \\
Quarot \small{\emph{w/} GPTQ. }& 5.48  &5.50  &6.22\\
Quarot-LLMC \small{\emph{w/} GPTQ-LLMC.} &5.48  & 5.50 &6.24\\
\bottomrule
\end{tabular}
}
\vspace{-0.1in}
\caption{Wikitext2 PPL alignment results of weight-activation asymmetric quantization of LLaMA-2-7B Model.}
\label{w_a_1}
\end{table}
\begin{table}[th!]
\centering
\resizebox{ \linewidth}{!}
{
\begin{tabular}{lcccc}
\toprule
Method & LLaMA-2-7b	 & LLaMA-2-70b & LLaMA-3-8b	& LLaMA-3-70b  \\
\midrule
Wanda & 6.91 & 4.22	 & 9.56 & OOM\\
Wanda-LLMC & 6.91 & 4.19 & 9.58 & 5.75 \\
\bottomrule
\end{tabular}
}
\vspace{-0.1in}
\caption{Wikitext2 PPL alignment results of 50\% unstructured sparsification method Wanda~\citep{sun2024simpleeffectivepruningapproach} for LLaMA-2-7B, 70B, and LLaMA-3 family.}
\label{w_a_2}
\end{table}
\begin{table}[th!]
\centering
\setlength{\tabcolsep}{6mm}
\resizebox{0.32\textwidth}{!}
{
\begin{tabular}{lc}
\toprule
Method & w8a8 \\
\midrule
SmoothQuant & 5.589 \\
SmoothQuant-LLMC & 5.589 \\
\cdashline{1-2}
OS+ & 5.511 \\
OS+-LLMC & 5.517 \\

\bottomrule
\end{tabular}
}
\vspace{-0.1in}
\caption{Wikitext2 PPL alignment results of weight-activation symmetric quantization of LLaMA-2-7B Model.}
\label{w_a_3}
\end{table}

  \begin{table}[th!]
    \renewcommand{\arraystretch}{1.7}
        \vspace{-0.1in}
        \centering
        \scalebox{0.5}{\begin{tabular}{ccccc}
\toprule
\multirow{2}{*}{\textbf{Model}} & \multirow{2}{*}{\textbf{KV Cache Prec.}} & \multicolumn{3}{c}{\textbf{Pass@1~(\%) $\uparrow$}} \\\cmidrule(l){3-5}
& & Human-Eval & MBPP & Avg. \\
\midrule
\multirow{5}{*}{LLaMA-2-7B} & Full Prec.     & 12.80               & 22.00         & 17.40         \\
\cdashline{2-5}
& int8  & 13.41               & 20.00         & 16.71         \\
& int4 & 13.41               & 21.00         & 17.21         \\
& int2 & 0.00               & 0.00         & 0.00         \\
& w4a8kv4 & 12.20               & 18.40         & 15.30         \\
\midrule
\multirow{5}{*}{LLaMA-2-13B} & Full Prec.   & 18.29               & 24.00         & 21.15       \\
\cdashline{2-5}
& int8 & 17.68               & 23.00         & 20.34         \\
& int4 & 17.68               & 23.00        & 20.34         \\
& int2 & 0.00               & 0.00         & 0.00         \\
& w4a8kv4 & 15.85               & 23.40         & 19.63         \\
\midrule
\multirow{5}{*}{LLaMA-2-70B} & Full Prec.    & 29.27               & 42.00        & 35.64         \\
\cdashline{2-5}
& int8 & 29.88               & 38.00         & 33.94         \\
& int4 & 30.49               & 39.00         & 34.75         \\
& int2 & 0.00               & 0.00         & 0.00         \\
& w4a8kv4 & 29.27               & 38.20         & 33.74         \\
\bottomrule
\end{tabular}
}
        \vspace{-0.1in}
        \caption{Naive KV cache quantization results on Human-Eval and MBPP for LLAMA-2 series models. We employ group-wise quantization~(\emph{i.e.}, g8) here.}
        \label{tab:kv-cache}
\end{table}
 \subsection{KV Cache Quantization}
 This part shows the accuracy of KV cache quantization for code generation tasks. From \autoref{tab:kv-cache}, we can find that the naive int8 and int4 KV cache quantization brings almost no accuracy degradation for both the Human-Eval and MBPP datasets. This conclusion proves that the naive 4-bit KV cache can be adopted without harm to performance. However, the naive 2-bit KV cache will bring a crash for the generation, and thus should not be adopted. Similar results can be found in \autoref{tab:kv-cache-long-context} for long-context evaluation.

 \subsection{Extensibility of LLMC}\label{fp-quantization}
To further demonstrate the extensibility of the toolkit, we conduct extensive experiments, including %
MoE quantization (shown in \autoref{tab:moe-quantization}), VLM quantization (shown in \autoref{tab:llava-quantization}), and sparsification (shown in \autoref{tab:wanda}).

\noindent\textbf{MOE quantization. } We utilize our toolkit to evaluate the performance of quantized Mixtral-8x7B, as shown in \autoref{tab:moe-quantization}.
 \begin{table}[!ht]
    \renewcommand{\arraystretch}{1.6}
    \vspace{-0.1in}
    \centering
    \scalebox{0.6}{\begin{tabular}{ccccccc}
\toprule
\multirow{2}{*}{\textbf{\#Bits}} & \multirow{2}{*}{\textbf{Method}} & \multicolumn{3}{c}{\textbf{PPL $\downarrow$}} \\\cmidrule(l){3-5} & & WikiText2 & C4 & Avg. \\
\midrule
Full Prec. & - & 3.84 & 7.40 &  5.62 \\
\cline{1-6}

\multirow{2}{*}{w4a16g128} & AWQ & 4.05 & 7.59 & 5.82 \\
& GPTQ& 4.05 & 7.60 & 5.82 \\
\midrule

\multirow{2}{*}{w3a16g128} & AWQ & 4.73 & 8.29 & 7.07 \\
& GPTQ& 4.93 & 8.52 & 7.18 \\
\midrule

\multirow{2}{*}{w8a8} & SmoothQuant & 3.87 & 7.48 & 5.68 \\
& OS+& 3.87 & 7.48 & 5.68 \\
\midrule

\multirow{2}{*}{w6a6} & SmoothQuant & 4.28 & 7.89 & 6.09 \\
& OS+& 4.27 & 7.90 & 6.09 \\

\bottomrule
\end{tabular}
}
    \vspace{-0.1in}
    \caption{Ablation results of Mixtral-8x7B weight-only quantization and weight-activation quantization.}
    \label{tab:moe-quantization}
\end{table}

\noindent\textbf{ VLM quantization. } For the VLM quantization, the quantized LLaVA-7B is evaluated by our toolkit on Perception and Cognition tasks, as depicted in \autoref{tab:llava-quantization}.
\begin{table}[!ht]
    \renewcommand{\arraystretch}{1.6}
    \vspace{-0.1in}
    \centering
    \scalebox{0.6}{\begin{tabular}{ccccccc}
\toprule
\multirow{2}{*}{\textbf{\#Bits}} & \multirow{2}{*}{\textbf{Method}} & \multicolumn{3}{c}{\textbf{PPL $\downarrow$}} \\\cmidrule(l){3-5} & & Perception & Cognition & Avg. \\
\midrule
Full Prec. & - & 1477.60 & 283.21 &  880.40 \\
\cline{1-6}

\multirow{2}{*}{w4a16g128} & AWQ & 1441.85 & 276.78 & 859.31 \\
& GPTQ& 1416.23 & 285.0  & 850.61 \\
\midrule

\multirow{2}{*}{w3a16g128} & AWQ & 1417.28 & 259.64 & 838.46 \\
& GPTQ& 1346.07 & 280.71 & 813.39 \\
\midrule

\multirow{2}{*}{w8a8} & SmoothQuant & 1468.93 & 281.07 &  875.0 \\
& OS+& 1467.28 & 280.71 & 873.99 \\
\midrule

\multirow{2}{*}{w6a6} & SmoothQuant & 1469.67 & 298.21 &  883.94 \\
& OS+& 1467.20 & 299.64 & 883.42 \\

\bottomrule
\end{tabular}
}
    \vspace{-0.1in}
    \caption{Ablation results of LLaVA-7B weight-only quantization and weight-activation quantization.}
    \label{tab:llava-quantization}
\end{table}

\noindent\textbf{ Sparsity. } \autoref{tab:wanda} presents the results for the LLaMA-2-7B, 70B, and LLaMA-3 family of models obtained using the sparsification method Wanda~\citet{sun2023simple}.

\noindent\textbf{Mixed precision.} \autoref{tab:mix_bits} presents the results for weight-only mixed precison on LLaMA-2-7B and LLaMA-3-8B. Mixed precision is an effective method for mitigating quantization errors. More than specific algorithms, LLMC also supports customized layer-wise bit allocation. We found that 5-bit to 8-bit precision for the \texttt{down\_proj} offer almost the same benefits.
\begin{table}[!ht]
    \renewcommand{\arraystretch}{1.6}
    \vspace{-0.1in}
    \centering
    \scalebox{0.6}{\begin{tabular}{ccc}
\hline
     & LLaMA-2-7B & LLaMA-3-8B \\ \hline
Full Prec.  & 5.47                    & 6.14                     \\
w3a16g128  & 6.16                     & 8.08                   \\
w3a16g128 \emph{w/} \texttt{down\_proj}-w8a16g128   & 5.93                     & 7.45                   \\
w3a16g128 \emph{w/} \texttt{down\_proj}-w6a16g128   & 5.94                     & 7.44                 \\
w3a16g128 \emph{w/} \texttt{down\_proj}-w5a16g128   & 5.95                    & 7.48                   \\
w3a16g128 \emph{w/} \texttt{down\_proj}-w4a16g128   & 5.99                    & 7.61                    \\ \hline
\end{tabular}
}
    \vspace{-0.1in}
    \caption{PPL results on Wikitext2 of mixed precision with AWQ. We only apply higher bit allocation for \texttt{down\_proj}, as it vastly impacts the performance mentioned in the main text.}
    \label{tab:mix_bits}
\end{table}

\subsection{Inference Speed}
\label{appendix:inference_speed}
\begin{figure}[th!]
    \centering
    \includegraphics[width=0.8\linewidth]{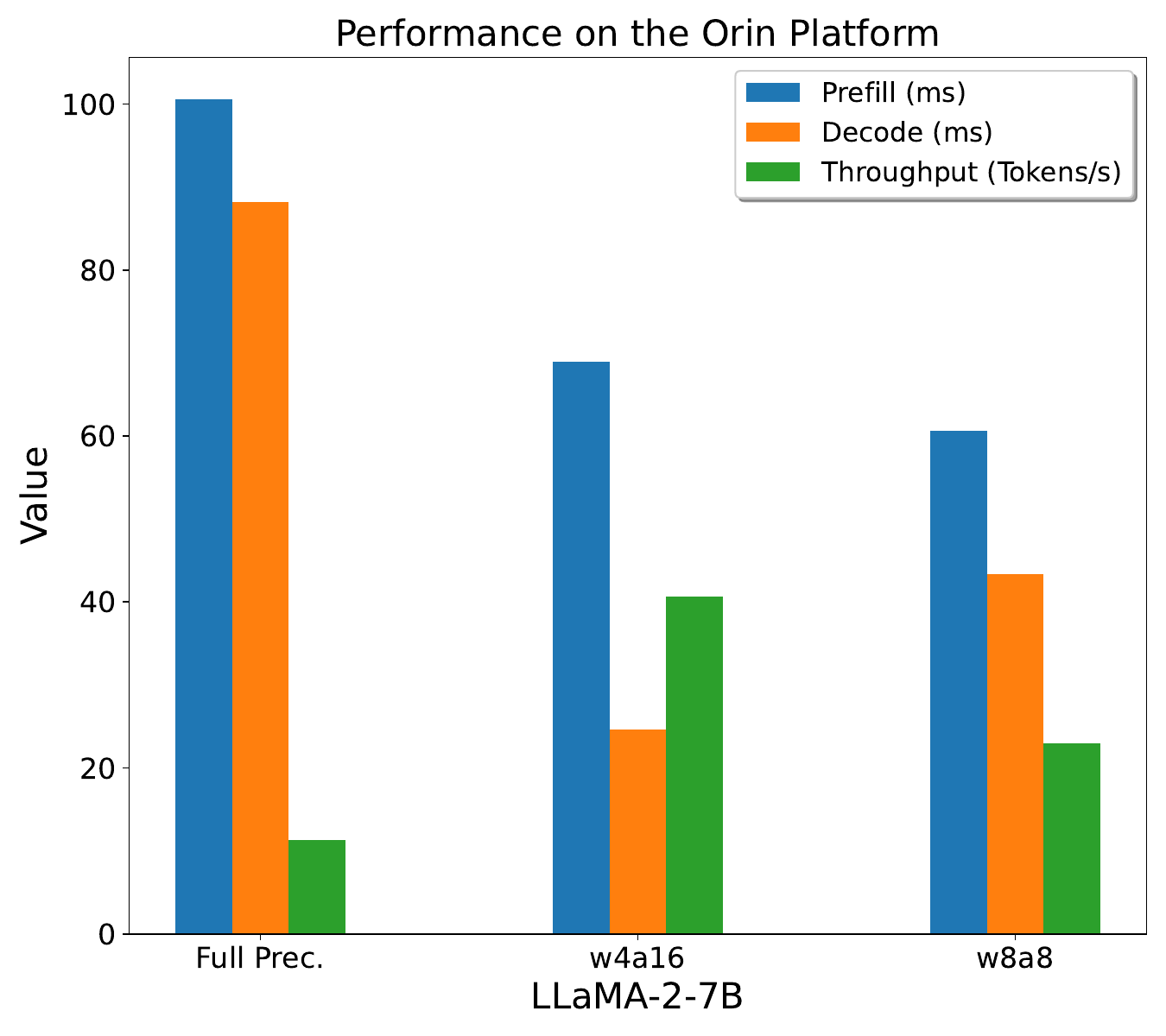}
    \vspace{-0.1in}
    \caption{Throughput comparison of quantization on the edge GPU (Drive Orin). (Token/s)} 
    \label{fig:speed-orin}
\end{figure}
\begin{figure*}[th!]
   \centering
       \includegraphics[width=0.7\linewidth]{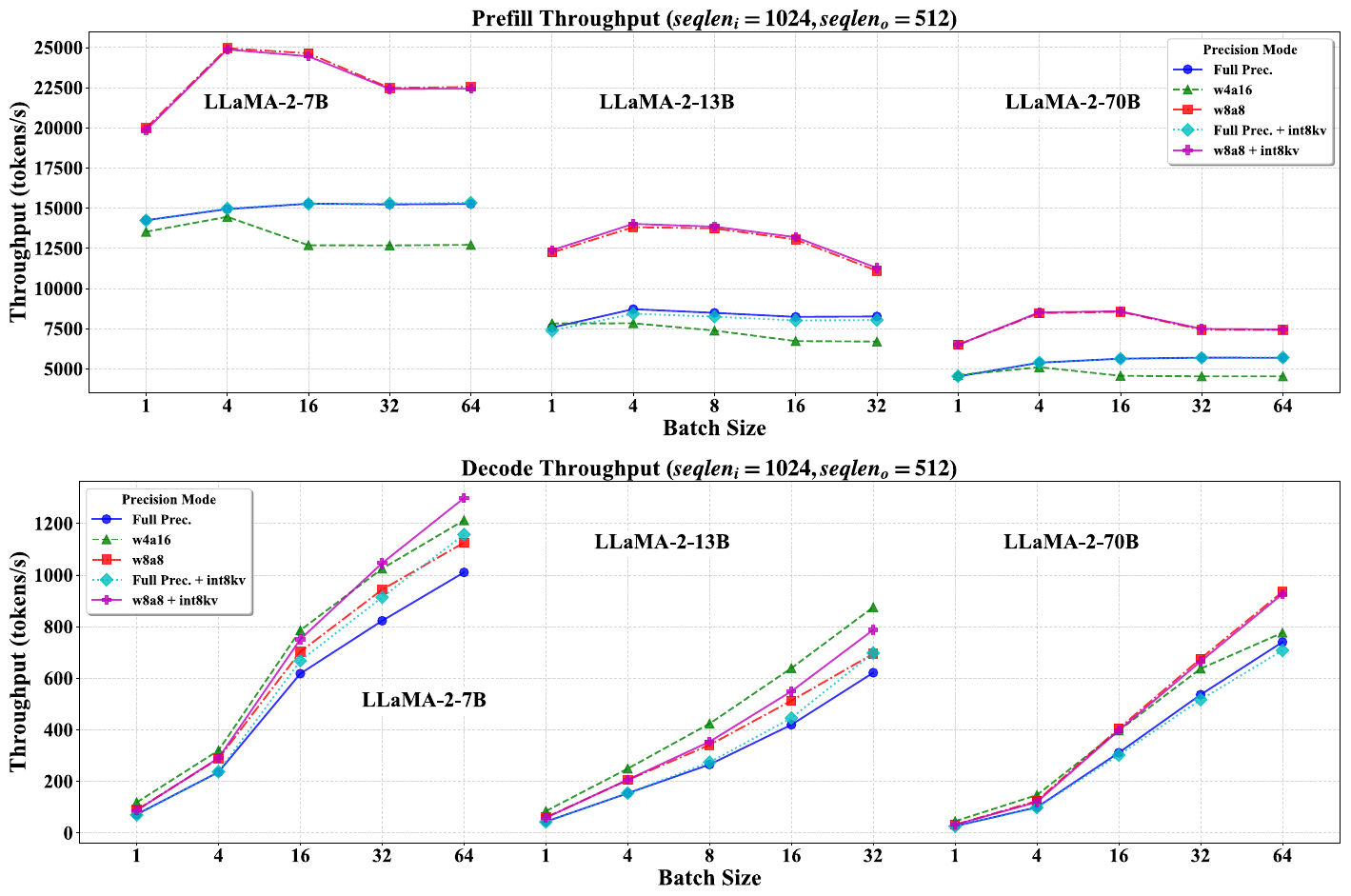}
        \vspace{-0.1in}
       \caption{Inference speed of 7B, 13B, and 70B LLaMA-2 models on NVIDIA A100 GPU. (Input sequence length: 1024, Output sequence length: 512)}
       \vspace{-0.1in}
   \label{fig:speed-1024-512}
\end{figure*}
\begin{figure*}[th!]
   \centering
       \includegraphics[width=0.7\linewidth]{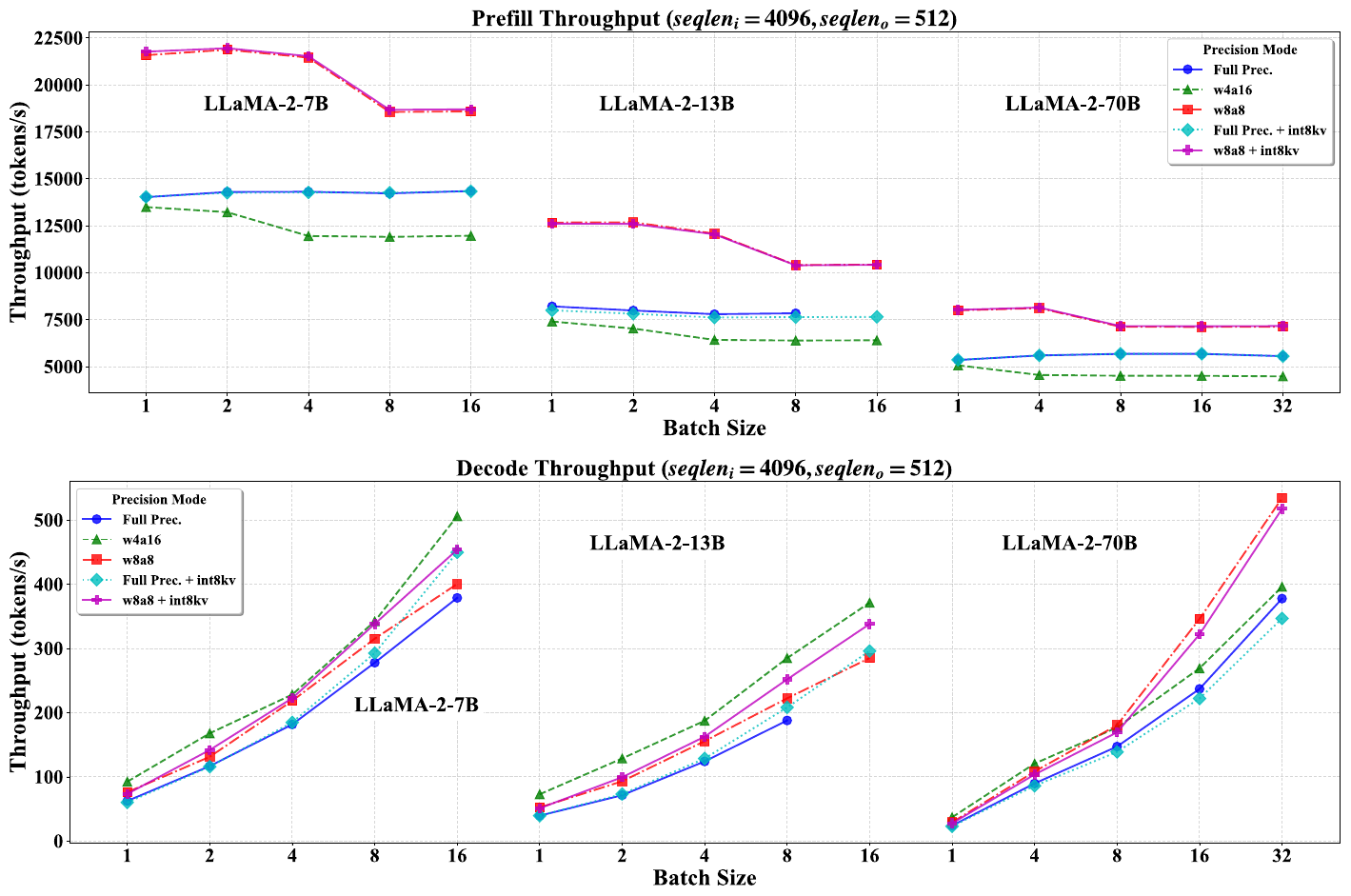}
        \vspace{-0.1in}
       \caption{Inference speed of 7B, 13B, and 70B LLaMA-2 models on NVIDIA A100 GPU. (Input sequence length: 4096, Output sequence length: 512)}
   \label{fig:speed-4096-512}
\end{figure*}
\begin{figure*}[th!]
   \centering
       \includegraphics[width=0.7\linewidth]{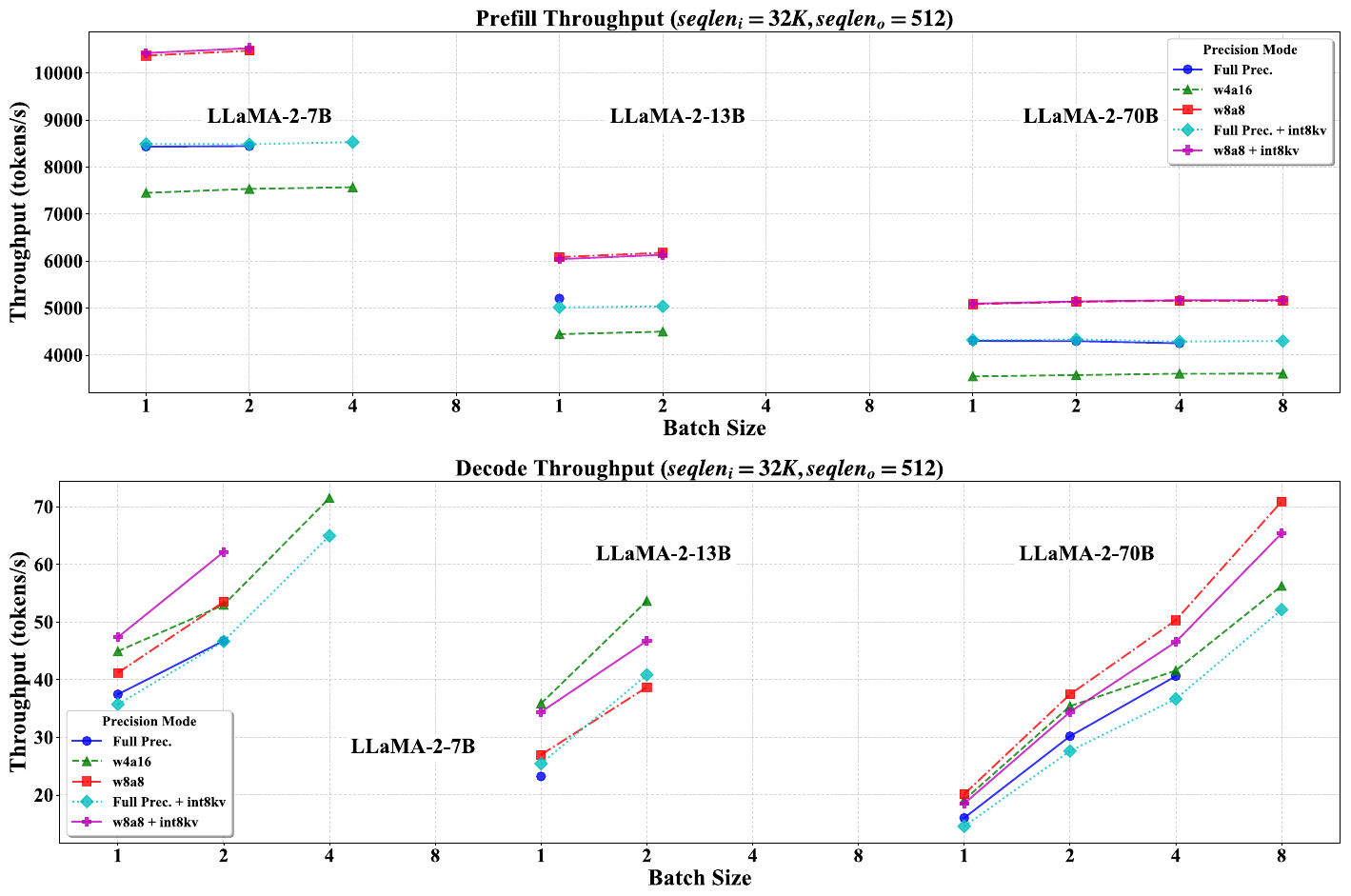}
        \vspace{-0.1in}
       \caption{Inference speed of 7B, 13B, and 70B LLaMA-2 models on NVIDIA A100 GPU. (Input sequence length: 32K, Output sequence length: 512)} %
   \label{fig:speed-32K-512}
\end{figure*}

To assess the practical benefits of different quantization approaches, we conducted evaluations~\footnote{In this section, all weight-only quantization employ 128g group-wise quantization.} using NVIDIA's cloud (SMX 80G A100) and edge (Drive Orin) GPUs, alongside the official inference library, TensorRT-LLM~\citep{trt-llm}. Part of our results, as depicted in \autoref{fig:speed-32K-512}, highlight the throughput improvements achieved for models with 32,000 input tokens and 512 output tokens. 
The findings indicate that quantization with 8-bit weights and activations enhances the prefill stage's speed by 20\%-30\% and the decode stage by 40\%-60\%. In contrast, 4-bit weight-only quantization reduces the prefill speed by 10\% but increases the decode speed by 40\%-60\%. It's important to note that these acceleration rates tend to diminish for larger models. Besides, 8-bit KV cache quantization has minimal impact on prefill times and slightly reduces decoding throughput for very large models, such as those with 70B models. \autoref{fig:speed-1024-512} and \autoref{fig:speed-4096-512} supplementarily illustrated the speedup brought by various quantization schemes on 1K and 4K input context length. We can also find that the conclusion for these two scenarios is the same as the 32K input context length. Moreover, \autoref{fig:speed-orin} shows the speed up on the Drive Orin edge GPU. It can be seen that weight-only quantization also helps the prefill under this setting, which is different from cloud GPUs.

\subsection{Detailed Accuracy \& PPL}
\label{detail}
\begin{table*}[th!]
    \renewcommand{\arraystretch}{1.2}
        \vspace{-0.1in}
        \centering
        \scalebox{0.6}{\begin{tabular}{cccccccc}
\toprule
\multirow{2}{*}{\textbf{Model}} & \multirow{2}{*}{\textbf{KV Cache Prec.}} & \multicolumn{5}{c}{\textbf{Accuracy~(\%) $\uparrow$}} \\\cmidrule(l){3-7}
                                &                                         & NarrativeQA & QASPER & MultiFieldQA-en & MultiFieldQA-zh & Avg. \\
\midrule
\multirow{4}{*}{ChatGLM3-6B-32k} & Full Prec.     & 25.93 & 43.35 & 51.57 & 62.36 & 45.80 \\
\cdashline{2-7}
                                 & int8  & 25.74 & 43.57 & 51.81 & 62.48 & 45.90 \\
                                 & int4  & 26.13 & 43.43 & 51.63 & 61.04 & 45.56 \\
                                 & int2  & 1.89 & 4.68 & 3.13 & 1.08 & 2.70 \\
\bottomrule
\end{tabular}
}
        \vspace{-0.1in}
        \caption{KV cache quantization results on Single-Document QA from LongBench~\cite{bai2023longbench}}
        \label{tab:kv-cache-long-context}  
\end{table*}
\begin{table*}[!ht]
    \setlength{\tabcolsep}{4mm}
    \centering
    \resizebox{ 0.7\linewidth}{!}{\begin{tabular}{ccccccccc}
\toprule

 \multirow{4}{*}{\textbf{Model}} & \multicolumn{8}{c}{\textbf{Sparsity}} \\ \cmidrule(l){2-3} \cmidrule(l){2-9} & \multicolumn{2}{c}{Dense} & \multicolumn{2}{c}{25$\%$} & \multicolumn{2}{c}{50$\%$} & \multicolumn{2}{c}{75$\%$}  \\ \cmidrule(l){2-3} \cmidrule(l){4-5} \cmidrule(l){6-7} \cmidrule(l){8-9} & C4 & Wikitext2 & C4 & Wikitext2 & C4 & Wikitext2 & C4 & Wikitext2  \\
\midrule
LLaMa2-7B	&7.26&	5.47&	7.46	&5.61&	9.25	&6.85&	260.42	&259.91\\
LLaMa2-70B	&5.71&	3.32	&5.76	&3.4&	6.49	&4.17	&32.5	&21.66\\
LLaMa3-8B	&9.44	&6.13&	10.01	&6.47&	15.07	&9.68&	336.62&	290.38\\
LLaMa3-70B	&7.16&	2.85	&7.44	&3.22&	9.96&	5.81	&93.99	&74.78\\
\bottomrule

\end{tabular}
}
    \vspace{-0.1in}
    \caption{Perplexity results of LLaMA-2-7B, 70B, and LLaMA-3 family under Wanda method.}
    \label{tab:wanda}
\end{table*}
 \begin{table*}[th!]
     \renewcommand{\arraystretch}{1.6}
     \vspace{-0.1in}
     \centering
     \scalebox{0.6}{\begin{tabular}{ccccc|cccccccc}
\toprule \multirow{2}{*}{\textbf{\#Bits}} & \multirow{2}{*}{\textbf{Method}} & \multicolumn{3}{c}{\textbf{PPL $\downarrow$}} & \multicolumn{6}{c}{\textbf{Accuracy (\%) $\uparrow$}} \\
\cmidrule(l){3-5} \cmidrule(l){6-13}
 & & WikiText2 & C4 & Avg. & MMLU & ARC-e & BoolQ & HellaSwag & PIQA & Avg. \\
\midrule

\multirow{2}{*}{w3a16g128} 

& AWQ  &6.22 &8.28 & 7.25 & 38.10 & 48.56 & 71.78 & 70.86 & 76.61 & 61.18 \\
\cdashline{2-13}

& AWQ \small{\emph{w/} asym. clip} &6.18 &8.24  & 7.21 & 42.33 & 47.09  & 71.44 & 70.93 &76.17 & 61.59 \\

\midrule
\multirow{2}{*}{w2a16g64} & AWQ  &2.09e5 &1.59e5 & 1.8e5 & 25.38& 4.87 &62.17 & 24.83 &51.2 & 37.69\\
\cdashline{2-13}
& AWQ \small{\emph{w/} asym. clip} &11.69 &14.83 & 13.26 & 27.4 & 25.4 &63.27 & 57.4 & 70.4& 48.77\\

\bottomrule
\end{tabular}
}
     \caption{Results of asymmetric/symmetric weight clipping for LLaMA-2-7B model.}
     \label{tab:7b-weight-only-detail}
 \end{table*}
\begin{table*}[th!]
     \renewcommand{\arraystretch}{1.6}
     \vspace{-0.1in}
     \centering
     \scalebox{0.6}{\begin{tabular}{ccccc|cccccccc}
\toprule \multirow{2}{*}{\textbf{\#Bits}} & \multirow{2}{*}{\textbf{Method}} & \multicolumn{3}{c}{\textbf{PPL $\downarrow$}} & \multicolumn{6}{c}{\textbf{Accuracy (\%) $\uparrow$}} \\
\cmidrule(l){3-5} \cmidrule(l){6-13}
 & & WikiText2 & C4 & Avg. & MMLU & ARC-e & BoolQ & HellaSwag & PIQA & Avg. \\
\midrule

\multirow{2}{*}{w3a16g128} 

& AWQ  & 3.75& 6.05 & 4.90  & 67.54& 87.65 &86.57 & 81.11 &81.88 & 80.95  \\
\cdashline{2-13}

& AWQ \small{\emph{w/} asym. clip} &3.74 &6.04 & 4.89 &67.07 & 89.95 & 86.30& 80.95 &81.07 & 81.07 \\

\midrule
\multirow{2}{*}{w2a16g64} & AWQ &7.1e4 &6.5e4 & 6.8e4 & 24.46 &26.46  &37.83 & 24.60 &50.87 & 32.84\\
\cdashline{2-13}
& AWQ \small{\emph{w/} asym. clip} &5.24 &7.73 & 6.49 & 57.91 & 80.07 &83.91 & 75.98 & 78.67& 75.31\\

\bottomrule
\end{tabular}
}
     \caption{Results of asymmetric/symmetric weight clipping for LLaMA-2-70B model.}
     \label{tab:70b-weight-only-detail}
 \end{table*}
 \begin{table*}[th!]
     \renewcommand{\arraystretch}{1.6}
     \vspace{-0.1in}
     \centering
     \scalebox{0.6}{\begin{tabular}{ccccc|cccccccc}
\toprule \multirow{2}{*}{\textbf{\#Bits}} & \multirow{2}{*}{\textbf{Method}} & \multicolumn{3}{c}{\textbf{PPL $\downarrow$}} & \multicolumn{6}{c}{\textbf{Accuracy (\%) $\uparrow$}} \\
\cmidrule(l){3-5} \cmidrule(l){6-13}
 & & WikiText2 & C4 & Avg. & MMLU & ARC-e & BoolQ & HellaSwag & PIQA & Avg. \\
\midrule

\multirow{4}{*}{w3a16g128} 

& GPTQ  &8.28 &13.07 &   10.67&57.81 &78.48 &73.49 &72.16 &77.86 &71.96\\
\cdashline{2-13}

& AWQ  &8.57 &13.39 &10.98 & 54.35 & 74.78 & 74.56 & 71.85 & 78.07 & 70.72\\
\cdashline{2-13}

& AWQ \small{\emph{w/} GPTQ} &8.18
&12.91  & 10.55& 59.10 & 80.60 &73.12  & 72.40 & 78.40 &72.72 \\
\cdashline{2-13}
& Quarot &40.81 &59.20 &50.00 & 29.03 & 29.98 & 58.87 & 45.18 & 66.43 & 45.90\\
\cdashline{2-13}
& Quarot \small{\emph{w/} GPTQ} &7.99
&12.70 &10.35 & 60.25 & 83.25 & 78.56 & 72.96 & 79.16 & 74.84\\

\bottomrule
\end{tabular}
}
     \caption{Results of reconstruction (GPTQ) combined with scaling (AWQ) and rotation-based (QuaRot) transformation for LLaMA-3-8B model. Clipping for AWQ here is canceled to expel distractions.}
     \label{tab:8b-weight-only-detail}
 \end{table*}
This section presents detailed data from some of the experiments discussed in the main text. \autoref{tab:7b-weight-only-detail} and \autoref{tab:70b-weight-only-detail} shows the detailed data for \autoref{tab:awq-clip}. \autoref{tab:8b-weight-only-detail} shows the detailed data for \autoref{tab:recon-trans}.

\subsection{Results for Various Model Families}\label{sec:scales}
From \autoref{tab:8b-weight-only-detail-0} to \autoref{tab:8b-weight-only-detail-6}, we report quantization results for different model families, including SmolLM~\footnote{\url{https://huggingface.co/blog/smollm}}, MiniCPM~\cite{hu2024minicpm}, and Qwen2~\cite{yang2024qwen2}. We additionally provide the results on SIQA~\cite{sap2019socialiqa}, ARC-c~\cite{allenai:arc}, OBQA~\cite{luo2021simple}, and WinoGrande~\cite{sakaguchi2019winogrande}.

\begin{table*}[th!]
     \renewcommand{\arraystretch}{1.6}
     \setlength{\tabcolsep}{4pt}
     \vspace{-0.1in}
     \centering
     \scalebox{0.6}{\begin{tabular}{lcccc|ccccccccc}
\toprule \multirow{2}{*}{\textbf{\#Bits}} & \multirow{2}{*}{\textbf{Method}} & \multicolumn{3}{c}{\textbf{PPL $\downarrow$}} & \multicolumn{9}{c}{\textbf{Accuracy (\%) $\uparrow$}} \\
\cmidrule(l){3-5} \cmidrule(l){6-14}
 & & WikiText2 & C4 & Avg. & ARC-e & ARC-c & BoolQ &  PIQA & SIQA & HellaS. & OBQA & WinoG. & Avg. \\
\midrule

Full Prec. & - & 17.56 & 22.17 & 19.86 & 53.04 & 39.51 & 68.34 & 23.00 & 35.14 & 60.00 & 61.36 & 25.94 & 45.79 \\
\cdashline{1-14}

\multirow{3}{*}{w2a16g128}&RTN & 2.27e+07 & 3.06e+07 & 2.66e+07 & 51.38 & 34.19 & \textbf{52.61} & \textbf{18.80} & 25.84 & \textbf{47.74} & 24.66 & \textbf{21.33} & \textbf{34.57} \\
&GPTQ & 1.30e+04 & 1.04e+04 & 1.17e+04 & 52.57 & 33.16 & 50.98 & 16.80 & \textbf{25.94} & 45.26 & \textbf{27.86} & 20.82 & 34.17 \\
&AWQ & \textbf{1.02e+04} & \textbf{8.18e+03} & \textbf{9.18e+03} & 48.93 & \textbf{34.44} & 51.03 & 15.60 & 25.62 & 38.84 & 26.30 & 20.48 & 32.65 \\
\cdashline{1-14}

\multirow{3}{*}{w3a16g128}&RTN & 91.65 & 96.75 & 94.20 & 48.38 & 36.59 & 60.88 & 19.00 & 30.21 & 49.54 & 46.84 & 21.93 & 39.17 \\
&GPTQ & \textbf{32.89} & \textbf{40.29} & \textbf{36.59} & 51.93 & 37.15 & 61.53 & \textbf{20.80} & \textbf{31.39} & \textbf{58.56} & \textbf{51.89} & 22.53 & \textbf{41.97} \\
&AWQ & 54.20 & 55.94 & 55.07 & 50.36 & \textbf{37.41} & \textbf{62.40} & 17.00 & 31.28 & 52.23 & 51.56 & \textbf{24.49} & 40.84 \\
\cdashline{1-14}

\multirow{3}{*}{w4a16g128}&RTN & 22.54 & 28.04 & 25.29 & 53.67 & 38.54 & 66.38 & \textbf{23.00} & \textbf{34.18} & \textbf{62.05} & 58.04 & \textbf{25.85} & \textbf{45.21} \\
&GPTQ & \textbf{20.03} & \textbf{25.01} & \textbf{22.52} & 52.01 & \textbf{39.71} & 65.56 & 22.00 & 33.99 & 56.36 & \textbf{58.84} & 24.74 & 44.15 \\
&AWQ & 21.42 & 26.19 & 23.81 & 52.25 & 38.02 & \textbf{66.76} & 22.80 & 34.07 & 58.96 & 58.54 & 25.77 & 44.65 \\
\cdashline{1-14}

\multirow{4}{*}{w4a4} & RTN & 2.60e+03 & 2.22e+03 & 2.41e+03 & 50.91 & 33.73 & 52.61 & 17.40 & 26.40 & 43.73 & 30.77 & 18.77 & 34.29 \\
& SmoothQuant & 331.70 & 441.95 & 386.82 & 52.09 & 33.32 & 53.70 & \textbf{18.20} & 27.38 & \textbf{44.46} & 37.42 & 20.65 & 35.90 \\
& OS+ & \textbf{263.76} & \textbf{389.67} & \textbf{326.71} & 52.49 & \textbf{35.62} & 55.39 & 14.60 & \textbf{27.56} & 43.46 & 41.46 & \textbf{20.73} & \textbf{36.41} \\
& QuaRot & 472.15 & 567.85 & 520.00 & 49.17 & 34.34 & \textbf{56.37} & 14.60 & 27.08 & 41.01 & \textbf{43.01} & \textbf{20.73} & 35.79 \\
\cdashline{1-14}

\multirow{4}{*}{w6a6} & RTN & 22.84 & 27.45 & 25.14 & 49.41 & 38.28 & 65.07 & 20.00 & 33.02 & 58.23 & 56.73 & \textbf{25.68} & 43.30 \\
& SmoothQuant & 20.37 & 25.12 & 22.74 & 53.91 & 38.13 & 64.64 & \textbf{22.80} & 32.52 & 59.02 & 59.22 & 25.00 & 44.41 \\
& OS+ & \textbf{19.67} & \textbf{25.00} & \textbf{22.33} & 51.54 & \textbf{39.71} & \textbf{66.81} & 21.20 & 32.88 & \textbf{59.85} & \textbf{60.19} & 24.32 & 44.56 \\
& QuaRot & 20.26 & 25.02 & 22.64 & 52.25 & 39.05 & 66.32 & 22.40 & \textbf{33.06} & 57.77 & 60.14 & \textbf{25.68} & \textbf{44.58} \\
\cdashline{1-14}

\multirow{4}{*}{w8a8} & RTN & 17.75 & 22.45 & 20.10 & 52.57 & 39.05 & \textbf{68.01} & 21.80 & 35.07 & \textbf{60.37} & 61.45 & 25.09 & 45.43 \\
& SmoothQuant & 17.68 & 22.35 & 20.01 & 52.64 & \textbf{39.66} & 67.74 & 21.80 & \textbf{35.14} & 60.15 & 61.49 & 25.43 & 45.51 \\
& OS+ & \textbf{17.67} & \textbf{22.32} & \textbf{19.99} & 53.51 & 39.00 & 67.79 & \textbf{23.00} & \textbf{35.14} & 60.09 & \textbf{61.66} & \textbf{25.85} & \textbf{45.76} \\
& QuaRot & 17.77 & 22.42 & 20.10 & 52.33 & 39.15 & \textbf{68.01} & 22.80 & \textbf{35.14} & 60.34 & 61.15 & 25.34 & 45.53 \\
\bottomrule
\end{tabular}
}
     \caption{Quantization Results for SmolLM-135M model. Activation clipping and online rotation within QuaRot are canceled for a fair comparison. ``HellaS.'' and ``WinoG.'' represent HellaSwag and WinoGrande, respectively. We mark the best results in \textbf{bold}.}
     \label{tab:8b-weight-only-detail-0}
 \end{table*}

 \begin{table*}[th!]
     \renewcommand{\arraystretch}{1.6}
     \setlength{\tabcolsep}{4pt}
     \vspace{-0.1in}
     \centering
     \scalebox{0.6}{\begin{tabular}{lcccc|ccccccccc}
\toprule \multirow{2}{*}{\textbf{\#Bits}} & \multirow{2}{*}{\textbf{Method}} & \multicolumn{3}{c}{\textbf{PPL $\downarrow$}} & \multicolumn{9}{c}{\textbf{Accuracy (\%) $\uparrow$}} \\
\cmidrule(l){3-5} \cmidrule(l){6-14}
 & & WikiText2 & C4 & Avg. & ARC-e & ARC-c & BoolQ &  PIQA & SIQA & HellaS. & OBQA & WinoG. & Avg. \\
\midrule
Full Prec. & - & 13.10 & 17.68 & 15.39 & 58.25 & 41.25 & 71.33 & 25.20 & 41.63 & 55.20 & 69.82 & 33.28 & 49.49 \\
\cdashline{1-14}

\multirow{3}{*}{w2a16g128}&RTN & 2.98e+06 & 2.60e+06 & 2.79e+06 & 51.22 & 32.91 & 51.74 & \textbf{16.40} & 25.72 & \textbf{47.95} & 25.21 & \textbf{20.90} & \textbf{34.01} \\
&GPTQ & \textbf{797.15} & \textbf{812.25} & \textbf{804.70} & 48.62 & \textbf{34.95} & 50.22 & 16.00 & 26.24 & 39.30 & 27.69 & 18.69 & 32.71 \\
&AWQ & 3.12e+03 & 2.67e+03 & 2.90e+03 & 48.93 & 34.08 & \textbf{52.50} & 15.20 & \textbf{26.82} & 42.11 & \textbf{30.68} & 19.97 & 33.79 \\
\cdashline{1-14}

\multirow{3}{*}{w3a16g128}&RTN & 32.13 & 39.52 & 35.83 & 53.35 & 36.80 & \textbf{67.30} & \textbf{22.20} & 36.23 & \textbf{62.02} & 57.87 & 29.44 & \textbf{45.65} \\
&GPTQ & \textbf{21.14} & \textbf{26.85} & \textbf{24.00} & 52.64 & 37.77 & 65.56 & 19.40 & 36.47 & 51.96 & 57.95 & 27.99 & 43.72 \\
&AWQ & 23.24 & 28.91 & 26.08 & 53.75 & \textbf{38.28} & 66.76 & 21.00 & \textbf{37.86} & 53.91 & \textbf{61.41} & \textbf{29.86} & 45.35 \\
\cdashline{1-14}

\multirow{3}{*}{w4a16g128}&RTN & 15.11 & 20.20 & 17.65 & 56.20 & 40.53 & \textbf{70.46} & \textbf{24.20} & 40.39 & 54.37 & 65.87 & 32.00 & 48.00 \\
&GPTQ & \textbf{14.80} & \textbf{19.72} & \textbf{17.26} & 55.72 & 39.36 & 69.91 & 23.80 & 39.75 & \textbf{54.43} & 66.20 & 31.14 & 47.54 \\
&AWQ & 15.17 & 20.08 & 17.63 & 57.06 & \textbf{40.94} & 69.26 & 23.00 & \textbf{41.00} & 51.74 & \textbf{68.27} & \textbf{32.85} & \textbf{48.02} \\
\cdashline{1-14}

\multirow{4}{*}{w4a4} & RTN & 645.64 & 613.99 & 629.82 & 51.14 & 33.88 & 54.52 & 13.80 & 26.51 & 43.61 & 33.96 & 19.37 & 34.60 \\
& SmoothQuant & 123.40 & 233.90 & 178.65 & 48.70 & 35.36 & \textbf{59.47} & \textbf{17.20} & 30.35 & 45.17 & 44.87 & \textbf{24.66} & \textbf{38.22} \\
& OS+ & \textbf{80.14} & \textbf{122.98} & \textbf{101.56} & 49.96 & \textbf{35.41} & 58.43 & 13.20 & \textbf{30.46} & \textbf{47.06} & \textbf{48.70} & 21.67 & 38.11 \\
& QuaRot & 157.89 & 158.13 & 158.01 & 49.41 & 34.44 & 57.73 & 15.80 & 28.28 & 39.08 & 40.57 & 21.16 & 35.81 \\
\cdashline{1-14}

\multirow{4}{*}{w6a6} & RTN & 15.32 & 21.15 & 18.24 & 55.17 & 40.23 & 69.15 & 23.00 & 39.44 & 48.78 & 66.46 & 30.89 & 46.64 \\
& SmoothQuant & 14.26 & 19.17 & 16.72 & 53.20 & 40.99 & 69.53 & \textbf{26.80} & 40.84 & 53.98 & 67.85 & 32.08 & \textbf{48.16} \\
& OS+ & \textbf{14.15} & \textbf{19.01} & \textbf{16.58} & 54.14 & \textbf{41.40} & \textbf{69.75} & 23.00 & \textbf{40.86} & 53.88 & 67.34 & \textbf{32.42} & 47.85 \\
& QuaRot & 14.36 & 19.24 & 16.80 & 54.30 & 40.84 & 69.64 & 24.40 & 40.41 & \textbf{55.05} & \textbf{68.35} & 32.00 & 48.12 \\
\cdashline{1-14}

\multirow{4}{*}{w8a8} & RTN & 13.31 & 17.97 & 15.64 & 56.04 & 40.58 & 70.67 & 25.80 & 41.64 & 55.20 & \textbf{70.24} & 33.79 & 49.24 \\
& SmoothQuant & 13.27 & 17.90 & 15.58 & 56.75 & \textbf{41.30} & 70.95 & 25.80 & 41.67 & \textbf{55.96} & 70.03 & 33.53 & 49.50 \\
& OS+ & \textbf{13.24} & \textbf{17.85} & \textbf{15.55} & 55.96 & 41.10 & \textbf{71.16} & \textbf{26.20} & 41.67 & 55.84 & 70.16 & \textbf{34.04} & \textbf{49.52} \\
& QuaRot & 13.26 & 17.90 & 15.58 & 56.75 & 40.89 & \textbf{71.16} & 25.20 & \textbf{41.73} & 53.82 & 69.87 & 33.87 & 49.16 \\

\bottomrule
\end{tabular}
}
     \caption{Quantization Results for SmolLM-350M model.}
     \label{tab:8b-weight-only-detail-1}
 \end{table*}

 \begin{table*}[th!]
     \renewcommand{\arraystretch}{1.6}
     \setlength{\tabcolsep}{4pt}
     \vspace{-0.1in}
     \centering
     \scalebox{0.6}{\begin{tabular}{lcccc|ccccccccc}
\toprule \multirow{2}{*}{\textbf{\#Bits}} & \multirow{2}{*}{\textbf{Method}} & \multicolumn{3}{c}{\textbf{PPL $\downarrow$}} & \multicolumn{9}{c}{\textbf{Accuracy (\%) $\uparrow$}} \\
\cmidrule(l){3-5} \cmidrule(l){6-14}
 & & WikiText2 & C4 & Avg. & ARC-e & ARC-c & BoolQ &  PIQA & SIQA & HellaS. & OBQA & WinoG. & Avg. \\
\midrule
Full Prec. & - & 9.58 & 13.92 & 11.75 & 60.93 & 43.65 & 75.79 & 30.00 & 49.55 & 65.93 & 76.47 & 43.43 & 55.72 \\
\cdashline{1-14}

\multirow{3}{*}{w2a16g128}&RTN & 1.40e+07 & 1.06e+07 & 1.23e+07 & 49.64 & 33.42 & 53.10 & \textbf{17.20} & 25.85 & 44.50 & 25.42 & 22.61 & 33.97 \\
&GPTQ & 465.98 & 319.93 & 392.95 & 51.70 & 34.60 & 51.25 & 15.60 & 27.03 & 51.38 & 30.68 & 19.28 & 35.19 \\
&AWQ & \textbf{91.93} & \textbf{122.20} & \textbf{107.06} & 49.64 & \textbf{34.65} & \textbf{60.72} & 16.40 & \textbf{31.11} & \textbf{56.36} & \textbf{50.38} & \textbf{23.38} & \textbf{40.33} \\
\cdashline{1-14}

\multirow{3}{*}{w3a16g128}&RTN & 17.57 & 23.43 & 20.50 & 56.99 & 41.20 & 72.36 & \textbf{28.60} & \textbf{45.72} & 61.47 & 70.20 & \textbf{39.93} & 52.06 \\
&GPTQ & \textbf{12.10} & 16.85 & 14.47 & 58.56 & 40.89 & 73.01 & 27.80 & 45.21 & 61.56 & 71.09 & 37.37 & 51.94 \\
&AWQ & 12.11 & \textbf{16.68} & \textbf{14.40} & 57.70 & \textbf{41.81} & \textbf{73.34} & 28.20 & 45.22 & \textbf{63.91} & \textbf{72.81} & 39.76 & \textbf{52.84} \\
\cdashline{1-14}

\multirow{3}{*}{w4a16g128}&RTN & 10.56 & 15.13 & 12.85 & 60.30 & \textbf{44.52} & 75.08 & \textbf{31.20} & \textbf{49.12} & 63.00 & \textbf{76.05} & \textbf{43.52} & \textbf{55.35} \\
&GPTQ & \textbf{10.05} & 14.45 & 12.25 & 60.54 & 43.76 & 74.97 & 29.40 & 48.43 & 65.29 & 75.67 & 42.41 & 55.06 \\
&AWQ & 10.05 & \textbf{14.43} & \textbf{12.24} & 60.77 & 43.50 & \textbf{75.79} & 29.60 & 48.56 & \textbf{65.57} & 75.97 & 42.92 & 55.34 \\
\cdashline{1-14}

\multirow{4}{*}{w4a4} & RTN & 1.34e+07 & 8.32e+07 & 4.83e+07 & 50.59 & 33.06 & 50.98 & 14.80 & 24.50 & 48.87 & 29.38 & 22.18 & 34.30 \\
& SmoothQuant & 285.34 & 222.59 & 253.96 & 51.62 & 34.24 & 54.46 & 15.60 & 29.47 & 55.78 & 42.68 & 23.29 & 38.39 \\
& OS+ & 403.41 & 882.42 & 642.91 & 47.99 & 36.03 & 55.77 & 17.40 & 29.64 & 54.04 & 47.60 & 25.00 & 39.18 \\
& QuaRot & \textbf{37.41} & \textbf{49.55} & \textbf{43.48} & 50.20 & \textbf{37.15} & \textbf{60.07} & \textbf{17.80} & \textbf{34.05} & \textbf{58.90} & \textbf{52.10} & \textbf{26.45} & \textbf{42.09} \\
\cdashline{1-14}

\multirow{4}{*}{w6a6} & RTN & 11.71 & 16.65 & 14.18 & 56.20 & 41.97 & 73.29 & 28.60 & 46.47 & 63.73 & 72.81 & 38.40 & 52.68 \\
& SmoothQuant & 10.71 & 15.54 & 13.12 & 59.35 & 42.27 & \textbf{74.43} & \textbf{30.40} & 47.87 & 64.46 & 74.37 & 39.85 & 54.12 \\
& OS+ & 10.51 & 15.13 & 12.82 & 58.96 & \textbf{42.43} & 73.99 & 29.20 & 48.25 & 64.83 & 73.78 & 40.44 & 53.98 \\
& QuaRot & \textbf{10.35} & \textbf{14.99} & \textbf{12.67} & 58.09 & \textbf{42.43} & 73.83 & 29.60 & \textbf{48.65} & \textbf{65.14} & \textbf{74.66} & \textbf{40.70} & \textbf{54.14} \\
\cdashline{1-14}

\multirow{4}{*}{w8a8} & RTN & 9.73 & 14.21 & 11.97 & 59.67 & \textbf{43.86} & \textbf{76.01} & \textbf{30.60} & 49.40 & 66.02 & 76.35 & 42.92 & 55.60 \\
& SmoothQuant & 9.65 & 14.04 & 11.84 & 61.33 & 43.50 & 75.63 & 30.40 & 49.37 & 65.81 & 76.60 & 43.00 & 55.71 \\
& OS+ & 9.64 & 14.01 & 11.83 & 60.46 & 43.65 & 75.63 & 30.00 & \textbf{49.43} & \textbf{66.36} & \textbf{76.73} & \textbf{44.03} & \textbf{55.79} \\
& QuaRot & \textbf{9.64} & \textbf{14.01} & \textbf{11.82} & 59.91 & 43.30 & 75.79 & 30.20 & 49.33 & \textbf{66.36} & 76.64 & 43.43 & 55.62 \\
\bottomrule
\end{tabular}
}
     \caption{Quantization Results for SmolLM-1.7B model.}
     \label{tab:8b-weight-only-detail-2}
 \end{table*}

 \begin{table*}[th!]
     \renewcommand{\arraystretch}{1.6}
     \setlength{\tabcolsep}{4pt}
     \vspace{-0.1in}
     \centering
     \scalebox{0.6}{\begin{tabular}{lcccc|ccccccccc}
\toprule \multirow{2}{*}{\textbf{\#Bits}} & \multirow{2}{*}{\textbf{Method}} & \multicolumn{3}{c}{\textbf{PPL $\downarrow$}} & \multicolumn{9}{c}{\textbf{Accuracy (\%) $\uparrow$}} \\
\cmidrule(l){3-5} \cmidrule(l){6-14}
 & & WikiText2 & C4 & Avg. & ARC-e & ARC-c & BoolQ &  PIQA & SIQA & HellaS. & OBQA & WinoG. & Avg. \\
\midrule
Full Prec. & - & 8.60 & 13.74 & 11.17 & 60.62 & 45.04 & 74.48 & 23.20 & 50.09 & 68.23 & 70.37 & 36.26 & 53.54 \\
\cdashline{1-14}

\multirow{3}{*}{w2a16g128}&RTN & 7.86e+03 & 1.61e+04 & 1.20e+04 & 50.91 & 34.54 & 53.10 & 13.80 & 25.90 & 40.70 & 26.39 & 22.44 & 33.47 \\
&GPTQ & \textbf{71.23} & \textbf{101.64} & \textbf{86.44} & 48.86 & 36.03 & 57.24 & 16.20 & 29.00 & \textbf{43.12} & 33.88 & 19.45 & 35.47 \\
&AWQ & 100.70 & 197.93 & 149.31 & 52.64 & \textbf{38.18} & \textbf{60.83} & \textbf{16.60} & \textbf{31.88} & 42.60 & \textbf{44.78} & \textbf{22.95} & \textbf{38.81} \\
\cdashline{1-14}

\multirow{3}{*}{w3a16g128}&RTN & 11.00 & 17.70 & 14.35 & 60.77 & 41.50 & 72.42 & 19.60 & 46.76 & 63.79 & 63.93 & 33.28 & 50.26 \\
&GPTQ & 10.34 & 16.44 & 13.39 & 60.62 & 42.99 & 71.60 & 21.80 & 46.40 & 60.64 & 65.11 & \textbf{35.24} & 50.55 \\
&AWQ & \textbf{10.01} & \textbf{16.23} & \textbf{13.12} & 59.67 & \textbf{44.52} & \textbf{72.63} & \textbf{22.40} & \textbf{47.07} & \textbf{65.38} & \textbf{66.84} & 34.30 & \textbf{51.60} \\
\cdashline{1-14}

\multirow{3}{*}{w4a16g128}&RTN & 8.98 & 14.35 & 11.67 & 59.43 & 44.37 & 73.07 & 23.20 & \textbf{49.42} & 67.13 & \textbf{69.40} & 36.35 & 52.80 \\
&GPTQ & 8.89 & \textbf{14.23} & \textbf{11.56} & 60.46 & 44.06 & \textbf{73.39} & 22.80 & 49.00 & 69.24 & 68.64 & 36.09 & 52.96 \\
&AWQ & \textbf{8.87} & 14.25 & 11.56 & 61.17 & \textbf{45.19} & 73.29 & \textbf{23.40} & 49.36 & \textbf{71.01} & 69.32 & \textbf{36.60} & \textbf{53.67} \\
\cdashline{1-14}

\multirow{4}{*}{w4a4} & RTN & 35.70 & 50.17 & 42.93 & 52.17 & 39.05 & 64.09 & 16.60 & 36.29 & 57.34 & 51.47 & 25.51 & 42.81 \\
& SmoothQuant & 19.75 & 30.51 & 25.13 & 52.33 & 40.48 & 65.45 & 19.00 & \textbf{40.68} & 60.40 & \textbf{55.18} & 28.07 & \textbf{45.20} \\
& OS+ & 21.72 & 33.72 & 27.72 & 51.22 & \textbf{40.79} & \textbf{65.67} & \textbf{20.20} & 40.59 & 59.82 & 53.79 & \textbf{28.67} & 45.09 \\
& QuaRot & \textbf{19.18} & \textbf{30.01} & \textbf{24.60} & 52.01 & 35.31 & 59.30 & 18.00 & 28.77 & \textbf{63.39} & 41.25 & 26.54 & 40.57 \\
\cdashline{1-14}

\multirow{4}{*}{w6a6} & RTN & 9.09 & 14.44 & 11.77 & 61.01 & \textbf{44.58} & \textbf{74.16} & 22.40 & 49.33 & 69.30 & 69.11 & \textbf{36.60} & \textbf{53.31} \\
& SmoothQuant & 9.03 & 14.39 & 11.71 & 60.06 & 44.11 & 73.18 & \textbf{23.40} & 49.25 & 69.24 & \textbf{69.70} & 36.09 & 53.13 \\
& OS+ & 9.05 & \textbf{14.38} & 11.72 & 59.83 & 44.52 & 73.88 & \textbf{23.40} & \textbf{49.48} & 68.93 & 68.94 & 36.01 & 53.12 \\
& QuaRot & \textbf{9.01} & 14.41 & \textbf{11.71} & 58.80 & 36.85 & 65.29 & 20.20 & 31.16 & \textbf{69.48} & 47.35 & 29.35 & 44.81 \\
\cdashline{1-14}

\multirow{4}{*}{w8a8} & RTN & 8.65 & 13.80 & 11.23 & 62.04 & 44.17 & \textbf{74.48} & \textbf{23.80} & 49.86 & 68.10 & 70.08 & \textbf{36.52} & \textbf{53.63} \\
& SmoothQuant & 8.64 & 13.79 & 11.21 & 59.91 & 44.11 & 74.32 & 22.20 & 49.90 & 68.32 & \textbf{70.29} & 35.67 & 53.09 \\
& OS+ & \textbf{8.63} & \textbf{13.78} & \textbf{11.21} & 59.91 & \textbf{44.63} & 74.16 & 23.00 & \textbf{49.92} & 68.17 & 70.08 & 35.67 & 53.19 \\
& QuaRot & 8.64 & 13.79 & 11.22 & 60.38 & 37.15 & 64.96 & 22.40 & 31.53 & \textbf{68.99} & 48.06 & 29.61 & 45.38 \\
\bottomrule
\end{tabular}
}
     \caption{Quantization Results for MiniCPM-1B model.}
     \label{tab:8b-weight-only-detail-3}
 \end{table*}

 \begin{table*}[th!]
     \renewcommand{\arraystretch}{1.6}
     \setlength{\tabcolsep}{4pt}
     \vspace{-0.1in}
     \centering
     \scalebox{0.6}{\begin{tabular}{lcccc|ccccccccc}
\toprule \multirow{2}{*}{\textbf{\#Bits}} & \multirow{2}{*}{\textbf{Method}} & \multicolumn{3}{c}{\textbf{PPL $\downarrow$}} & \multicolumn{9}{c}{\textbf{Accuracy (\%) $\uparrow$}} \\
\cmidrule(l){3-5} \cmidrule(l){6-14}
 & & WikiText2 & C4 & Avg. & ARC-e & ARC-c & BoolQ &  PIQA & SIQA & HellaS. & OBQA & WinoG. & Avg. \\
\midrule
Full Prec. & - & 8.16 & 13.00 & 10.58 & 63.14 & 47.24 & 76.22 & 28.60 & 52.88 & 73.58 & 74.66 & 42.58 & 57.36 \\
\cdashline{1-14}

\multirow{3}{*}{w2a16g128}&RTN & 612.79 & 880.31 & 746.55 & 49.01 & 35.52 & 56.64 & 15.80 & 28.51 & 58.93 & 31.86 & 20.14 & 37.05 \\
&GPTQ & 29.60 & 45.30 & 37.45 & 47.75 & 36.44 & 60.88 & 15.20 & 32.90 & 55.87 & 38.85 & 21.67 & 38.69 \\
&AWQ & \textbf{24.28} & \textbf{36.25} & \textbf{30.26} & 55.09 & \textbf{40.07} & \textbf{66.10} & \textbf{16.80} & \textbf{39.54} & \textbf{63.70} & \textbf{55.89} & \textbf{29.35} & \textbf{45.82} \\
\cdashline{1-14}

\multirow{3}{*}{w3a16g128}&RTN & 9.79 & 15.54 & 12.66 & 60.22 & 44.58 & \textbf{74.48} & 25.80 & 50.42 & 71.83 & 70.12 & \textbf{40.78} & 54.78 \\
&GPTQ & 9.56 & 15.29 & 12.43 & 61.33 & 43.91 & 73.50 & 25.40 & 50.23 & 73.12 & 69.74 & 37.97 & 54.40 \\
&AWQ & \textbf{9.18} & \textbf{14.68} & \textbf{11.93} & 60.85 & \textbf{46.21} & 74.05 & \textbf{27.20} & \textbf{51.07} & \textbf{73.36} & \textbf{71.76} & 40.27 & \textbf{55.60} \\
\cdashline{1-14}

\multirow{3}{*}{w4a16g128}&RTN & 8.40 & 13.43 & 10.92 & 64.96 & 47.34 & \textbf{76.22} & \textbf{28.80} & 52.77 & 73.70 & 74.45 & \textbf{42.24} & \textbf{57.56} \\
&GPTQ & 8.50 & 13.59 & 11.04 & 61.88 & \textbf{47.39} & 75.30 & 27.40 & 52.65 & \textbf{75.14} & 73.40 & 41.89 & 56.88 \\
&AWQ & \textbf{8.32} & \textbf{13.39} & \textbf{10.85} & 61.33 & 46.88 & 75.73 & \textbf{28.80} & \textbf{52.80} & 74.65 & \textbf{74.96} & 41.72 & 57.11 \\
\cdashline{1-14}

\multirow{4}{*}{w4a4} & RTN & 33.64 & 52.72 & 43.18 & 53.35 & 38.64 & 64.85 & 18.00 & 37.21 & 62.60 & 52.23 & 26.71 & 44.20 \\
& SmoothQuant & 17.20 & \textbf{28.01} & \textbf{22.61} & 53.99 & \textbf{41.91} & 68.39 & \textbf{23.20} & 42.71 & \textbf{63.88} & \textbf{60.02} & \textbf{33.28} & \textbf{48.42} \\
& OS+ & \textbf{17.15} & 28.22 & 22.68 & 53.75 & 41.15 & \textbf{68.50} & 20.40 & \textbf{43.25} & 63.82 & 59.22 & 31.91 & 47.75 \\
& QuaRot & 19.87 & 31.97 & 25.92 & 53.51 & 35.98 & 61.04 & 16.80 & 27.36 & 61.50 & 40.91 & 25.09 & 40.27 \\
\cdashline{1-14}

\multirow{4}{*}{w6a6} & RTN & 8.46 & 13.58 & 11.02 & 63.14 & \textbf{45.96} & 75.03 & 27.60 & \textbf{52.21} & 73.21 & 73.78 & 40.61 & 56.44 \\
& SmoothQuant & \textbf{8.43} & \textbf{13.49} & \textbf{10.96} & 62.59 & 45.24 & \textbf{75.63} & 27.80 & 52.03 & 72.75 & 73.99 & 41.21 & 56.41 \\
& OS+ & 8.45 & 13.51 & 10.98 & 61.33 & 45.55 & 74.65 & \textbf{28.00} & 52.01 & \textbf{74.07} & \textbf{74.16} & \textbf{41.81} & \textbf{56.45} \\
& QuaRot & 8.48 & 13.55 & 11.01 & 61.56 & 38.33 & 65.18 & 20.60 & 28.49 & 72.23 & 52.36 & 31.91 & 46.33 \\
\cdashline{1-14}

\multirow{4}{*}{w8a8} & RTN & \textbf{8.13} & 13.04 & \textbf{10.59} & 63.77 & 46.57 & \textbf{76.39} & 29.40 & \textbf{52.97} & 73.94 & \textbf{74.66} & 42.24 & \textbf{57.49} \\
& SmoothQuant & 8.17 & 13.04 & 10.60 & 63.06 & 46.93 & 76.33 & 29.20 & 52.80 & 73.73 & 74.62 & \textbf{42.32} & 57.37 \\
& OS+ & 8.18 & \textbf{13.04} & 10.61 & 63.06 & \textbf{47.19} & 76.17 & \textbf{29.60} & 52.80 & \textbf{74.01} & 74.28 & 41.89 & 57.38 \\
& QuaRot & 8.18 & 13.04 & 10.61 & 62.75 & 38.43 & 66.05 & 22.40 & 28.57 & 73.58 & 53.07 & 31.83 & 47.09 \\

\bottomrule
\end{tabular}
}
     \caption{Quantization Results for MiniCPM-2B model.}
     \label{tab:8b-weight-only-detail-4}
 \end{table*}
  \begin{table*}[th!]
     \renewcommand{\arraystretch}{1.6}
     \setlength{\tabcolsep}{4pt}
     \vspace{-0.1in}
     \centering
     \scalebox{0.6}{\begin{tabular}{lcccc|ccccccccc}
\toprule \multirow{2}{*}{\textbf{\#Bits}} & \multirow{2}{*}{\textbf{Method}} & \multicolumn{3}{c}{\textbf{PPL $\downarrow$}} & \multicolumn{9}{c}{\textbf{Accuracy (\%) $\uparrow$}} \\
\cmidrule(l){3-5} \cmidrule(l){6-14}
 & & WikiText2 & C4 & Avg. & ARC-e & ARC-c & BoolQ &  PIQA & SIQA & HellaS. & OBQA & WinoG. & Avg. \\
\midrule
Full Prec. & - & 13.58 & 18.97 & 16.27 & 57.70 & 43.04 & 69.48 & 21.40 & 38.35 & 61.04 & 54.76 & 25.51 & 46.41 \\
\cdashline{1-14}

\multirow{3}{*}{w2a16g128}&RTN & 2.09e+05 & 1.97e+05 & 2.03e+05 & 51.30 & 34.03 & 53.37 & 14.40 & 25.48 & 44.86 & 25.00 & \textbf{22.70} & 33.89 \\
&GPTQ & \textbf{1.34e+03} & \textbf{1.39e+03} & \textbf{1.37e+03} & 50.04 & \textbf{34.49} & \textbf{53.70} & 13.80 & 25.95 & 44.28 & 27.86 & 20.90 & 33.88 \\
&AWQ & 9.73e+03 & 8.82e+03 & 9.27e+03 & 48.54 & 33.06 & 53.37 & \textbf{15.00} & \textbf{26.30} & \textbf{46.36} & \textbf{28.75} & 20.14 & \textbf{33.94} \\
\cdashline{1-14}

\multirow{3}{*}{w3a16g128}&RTN & 32.82 & 45.18 & 39.00 & 52.64 & 37.51 & 62.62 & 18.80 & 33.34 & 45.08 & 46.04 & 23.38 & 39.93 \\
&GPTQ & \textbf{19.62} & \textbf{28.06} & \textbf{23.84} & 52.96 & 37.10 & \textbf{66.43} & \textbf{19.00} & 34.71 & \textbf{59.60} & \textbf{51.98} & \textbf{24.49} & \textbf{43.28} \\
&AWQ & 22.72 & 30.28 & 26.50 & 52.96 & \textbf{39.00} & 66.16 & 18.00 & \textbf{35.18} & 57.34 & 49.28 & 23.72 & 42.70 \\
\cdashline{1-14}

\multirow{3}{*}{w4a16g128}&RTN & 15.75 & 21.90 & 18.83 & 54.54 & 40.69 & 67.68 & \textbf{21.20} & 37.42 & \textbf{62.32} & 51.01 & 23.98 & 44.86 \\
&GPTQ & \textbf{14.86} & \textbf{20.80} & \textbf{17.83} & 55.49 & 41.04 & 67.90 & 21.00 & 37.47 & 59.17 & \textbf{56.86} & 24.57 & \textbf{45.44} \\
&AWQ & 14.90 & 20.86 & 17.88 & 56.99 & \textbf{41.20} & \textbf{68.44} & 19.20 & \textbf{37.50} & 59.45 & 52.90 & \textbf{24.83} & 45.06 \\
\cdashline{1-14}

\multirow{4}{*}{w4a4} & RTN & 1.09e+03 & 1.01e+03 & 1.05e+03 & 48.86 & 34.60 & 52.45 & 13.20 & 26.23 & 41.71 & 27.86 & 19.62 & 33.07 \\
& SmoothQuant & 172.65 & 232.83 & 202.74 & 49.72 & 34.49 & 54.73 & 12.80 & 27.93 & 45.66 & 32.58 & 21.33 & 34.90 \\
& OS+ & 261.88 & 271.76 & 266.82 & 52.09 & 33.93 & 56.75 & 15.20 & 28.44 & 46.02 & 33.84 & 21.33 & 35.95 \\
& QuaRot & \textbf{57.48} & \textbf{78.85} & \textbf{68.16} & 51.54 & \textbf{35.21} & \textbf{59.63} & \textbf{15.80} & \textbf{30.25} & \textbf{48.20} & \textbf{38.97} & \textbf{21.42} & \textbf{37.63} \\
\cdashline{1-14}

\multirow{4}{*}{w6a6} & RTN & 15.79 & 21.99 & 18.89 & 53.99 & 40.33 & 67.08 & 21.00 & 37.14 & 50.46 & 53.66 & 26.37 & 43.75 \\
& SmoothQuant & 15.29 & 21.25 & 18.27 & 55.09 & 41.04 & 67.19 & \textbf{21.40} & 37.63 & 54.34 & 53.37 & \textbf{26.54} & 44.58 \\
& OS+ & 15.32 & 21.22 & 18.27 & 54.78 & \textbf{42.07} & \textbf{68.44} & 20.40 & \textbf{37.92} & 53.33 & 54.76 & 25.85 & 44.69 \\
& QuaRot & \textbf{14.93} & \textbf{20.82} & \textbf{17.87} & 55.17 & 41.56 & 67.63 & \textbf{21.40} & 37.62 & \textbf{57.40} & \textbf{55.72} & 25.43 & \textbf{45.24} \\
\cdashline{1-14}

\multirow{4}{*}{w8a8} & RTN & 13.85 & 19.37 & 16.61 & 56.12 & 42.22 & 69.37 & \textbf{21.80} & \textbf{38.32} & 58.93 & 54.59 & 25.17 & 45.81 \\
& SmoothQuant & 13.72 & 19.20 & 16.46 & 56.99 & 42.37 & \textbf{69.80} & 21.00 & 38.29 & 59.97 & 54.71 & 25.60 & 46.09 \\
& OS+ & \textbf{13.70} & \textbf{19.16} & \textbf{16.43} & 58.33 & \textbf{42.73} & \textbf{69.80} & 21.20 & 38.29 & 59.79 & \textbf{55.51} & \textbf{25.85} & \textbf{46.44} \\
& QuaRot & 13.70 & 19.17 & 16.44 & 55.88 & 42.48 & 69.64 & \textbf{21.80} & 38.26 & \textbf{60.61} & 55.22 & 25.26 & 46.14 \\
\bottomrule
\end{tabular}
}
     \caption{Quantization Results for Qwen2-0.5B model.}
     \label{tab:8b-weight-only-detail-5}
 \end{table*}

 \begin{table*}[th!]
     \renewcommand{\arraystretch}{1.6}
     \setlength{\tabcolsep}{4pt}
     \vspace{-0.1in}
     \centering
     \scalebox{0.6}{\begin{tabular}{lcccc|ccccccccc}
\toprule \multirow{2}{*}{\textbf{\#Bits}} & \multirow{2}{*}{\textbf{Method}} & \multicolumn{3}{c}{\textbf{PPL $\downarrow$}} & \multicolumn{9}{c}{\textbf{Accuracy (\%) $\uparrow$}} \\
\cmidrule(l){3-5} \cmidrule(l){6-14}
 & & WikiText2 & C4 & Avg. & ARC-e & ARC-c & BoolQ &  PIQA & SIQA & HellaS. & OBQA & WinoG. & Avg. \\
\midrule
Full Prec. & - & 9.84 & 14.36 & 12.10 & 64.72 & 46.11 & 75.57 & 26.80 & 48.31 & 71.96 & 65.87 & 33.45 & 54.10 \\
\cdashline{1-14}

\multirow{3}{*}{w2a16g128}&RTN & 3.41e+04 & 2.45e+04 & 2.93e+04 & 50.20 & 32.75 & 50.60 & \textbf{15.00} & 25.84 & \textbf{46.12} & 25.29 & 20.48 & 33.28 \\
&GPTQ & 482.42 & 462.96 & 472.69 & 52.49 & 34.08 & 54.30 & 14.60 & 26.58 & 44.40 & 28.96 & \textbf{20.82} & 34.53 \\
&AWQ & \textbf{326.04} & \textbf{398.14} & \textbf{362.09} & 51.38 & \textbf{34.95} & \textbf{55.98} & 14.80 & \textbf{28.12} & 42.72 & \textbf{34.72} & 20.31 & \textbf{35.37} \\
\cdashline{1-14}

\multirow{3}{*}{w3a16g128}&RTN & 15.24 & 21.27 & 18.26 & 61.72 & 43.19 & 70.95 & 23.00 & 43.65 & 68.01 & 60.14 & \textbf{32.00} & 50.33 \\
&GPTQ & \textbf{12.39} & \textbf{18.54} & \textbf{15.47} & 61.25 & 43.24 & \textbf{71.98} & \textbf{25.20} & \textbf{44.39} & \textbf{68.44} & \textbf{62.50} & 30.89 & \textbf{50.99} \\
&AWQ & 13.47 & 19.40 & 16.43 & 62.04 & \textbf{43.45} & 71.22 & 23.20 & 44.11 & 65.96 & 59.55 & 28.07 & 49.70 \\
\cdashline{1-14}

\multirow{3}{*}{w4a16g128}&RTN & 10.59 & 15.29 & 12.94 & 64.01 & 44.73 & 74.59 & \textbf{26.40} & 47.21 & \textbf{72.39} & 62.46 & 31.48 & 52.91 \\
&GPTQ & \textbf{10.28} & \textbf{15.02} & \textbf{12.65} & 66.14 & 45.29 & 74.54 & \textbf{26.40} & \textbf{47.68} & 71.07 & 65.07 & \textbf{32.68} & 53.61 \\
&AWQ & 10.41 & 15.16 & 12.79 & 66.69 & \textbf{46.57} & \textbf{75.24} & 26.00 & 47.22 & 70.55 & \textbf{65.40} & 31.83 & \textbf{53.69} \\
\cdashline{1-14}

\multirow{4}{*}{w4a4} & RTN & 275.87 & 265.84 & 270.85 & 50.99 & 34.54 & 55.77 & 13.60 & 28.63 & 44.68 & 31.48 & 20.56 & 35.03 \\
& SmoothQuant & 85.82 & 105.29 & 95.56 & 48.93 & 35.16 & \textbf{59.52} & 16.60 & 32.30 & 45.60 & 37.29 & \textbf{23.98} & 37.42 \\
& OS+ & 98.76 & 115.03 & 106.89 & 50.67 & \textbf{37.10} & 56.96 & 13.00 & 31.65 & 46.79 & 36.41 & 21.42 & 36.75 \\
& QuaRot & \textbf{42.19} & \textbf{56.01} & \textbf{49.10} & 52.17 & 35.82 & 58.65 & \textbf{17.80} & \textbf{34.72} & \textbf{50.86} & \textbf{38.38} & 21.50 & \textbf{38.74} \\
\cdashline{1-14}

\multirow{4}{*}{w6a6} & RTN & 11.02 & 15.83 & 13.42 & 63.93 & 43.91 & 72.80 & 25.80 & 46.91 & 63.64 & 62.88 & 31.83 & 51.46 \\
& SmoothQuant & 10.94 & 15.74 & 13.34 & 63.30 & 44.83 & 73.12 & 25.80 & \textbf{47.41} & 65.57 & 63.80 & 32.42 & 52.03 \\
& OS+ & \textbf{10.84} & \textbf{15.59} & \textbf{13.22} & 63.77 & 45.14 & 73.18 & \textbf{27.40} & 47.27 & 62.35 & 62.25 & 32.25 & 51.70 \\
& QuaRot & 10.86 & 15.61 & 13.24 & 64.17 & \textbf{46.37} & \textbf{74.21} & 26.60 & 47.21 & \textbf{67.52} & \textbf{65.28} & \textbf{34.13} & \textbf{53.19} \\
\cdashline{1-14}

\multirow{4}{*}{w8a8} & RTN & 9.96 & 14.43 & 12.19 & 64.88 & 46.37 & 75.30 & 26.80 & \textbf{48.13} & 72.26 & 65.87 & 33.11 & 54.09 \\
& SmoothQuant & 9.97 & 14.41 & 12.19 & 65.67 & \textbf{47.13} & \textbf{75.35} & \textbf{27.40} & 47.98 & 72.20 & \textbf{67.34} & \textbf{33.19} & \textbf{54.53} \\
& OS+ & 9.93 & 14.31 & 12.12 & 65.82 & 46.88 & \textbf{75.35} & 26.40 & \textbf{48.13} & \textbf{72.42} & 65.53 & \textbf{33.19} & 54.22 \\
& QuaRot & \textbf{9.89} & \textbf{14.31} & \textbf{12.10} & 65.59 & 46.06 & 75.03 & 26.60 & 48.09 & 71.65 & 65.87 & 33.02 & 53.99 \\

\bottomrule
\end{tabular}
}
     \caption{Quantization Results for Qwen2-1.5B model.}
     \label{tab:8b-weight-only-detail-6}
 \end{table*}

\label{sec:appendix}

\end{document}